\documentclass[runningheads]{llncs}
\usepackage[table,xcdraw]{xcolor}
\usepackage{multirow}
\usepackage{longtable}
\definecolor{orange}{HTML}{F9CB9C}
\definecolor{red}{HTML}{EA9999}
\definecolor{yellow}{HTML}{FFFC9E}
\newcommand{\shadeText}[2]{\colorbox{#1}{#2}}

\usepackage[mobile]{eccv}

\usepackage{eccvabbrv}

\usepackage{graphicx}
\usepackage{booktabs}

\usepackage[accsupp]{axessibility}  

\usepackage{hyperref}

\usepackage{orcidlink}

\begin{document}

\title{DefenseSplat: Enhancing the Robustness of 3D Gaussian Splatting via Frequency-Aware Filtering} 

\titlerunning{DefenseSplat}

\author{Yiran Qiao\orcidlink{0009-0009-7777-7356} \and
Yiren Lu\orcidlink{0000-0002-5411-0411} \and
Yunlai Zhou \and
Rui Yang \and
Linlin Hou \and
Yu Yin\orcidlink{0000-0002-9588-5854} 
\and
Jing Ma\orcidlink{0000-0003-4237-6607}
}

\authorrunning{Y.~Qiao et al.}

\institute{Case Western Reserve University, Cleveland OH 44106, USA \\
\email{\{yxq350, yxl3538, yxz3057, rxy337, lxh633, yxy1421, jxm1384\}@case.edu}\\
Code: \url{https://github.com/yrqiao/DefenseSplat}}

\maketitle

\begin{figure}
    \centering
    \includegraphics[width=\textwidth]{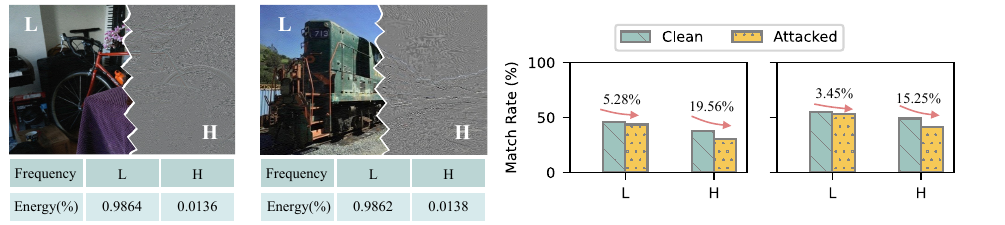}
    \caption{Effects of Attacks for 3DGS.  \textbf{Left}: Two examples showing the comparison between low- and high-frequency components on the same image after attack, with energy ratios of different frequency components in clean images at the bottom. The left side shows results on the \textit{Mip-NeRF 360} dataset, while the right side shows \textit{Tanks-and-Temples}. We take the \textit{bonsai} and \textit{Train} as examples, respectively. \textbf{Right}: Matching Rate (a signal for multi-view consistency) in low-frequency (L) and high-frequency components (H) of the clean/attacked images.}
    \label{fig1}
\end{figure}

\begin{abstract}
  3D Gaussian Splatting (3DGS) has emerged as a powerful paradigm for real-time and high-fidelity 3D reconstruction from posed images. However, recent studies reveal its vulnerability to adversarial corruptions in input views, where imperceptible yet consistent perturbations can drastically degrade rendering quality, increase training and rendering time, and inflate memory usage, even leading to server denial-of-service. In our work, to mitigate this issue, we begin by analyzing the distinct behaviors of adversarial perturbations in the low- and high-frequency components of input images using wavelet transforms. Based on this observation, we design a simple yet effective frequency-aware defense strategy that reconstructs training views by filtering high-frequency noise while preserving low-frequency content. This approach effectively suppresses adversarial artifacts while maintaining the authenticity of the original scene. Notably, it does not significantly impair training on clean data, achieving a desirable trade-off between robustness and performance on clean inputs. Through extensive experiments under a wide range of attack intensities on multiple benchmarks, we demonstrate that our method substantially enhances the robustness of 3DGS without access to clean ground-truth supervision. By highlighting and addressing the overlooked vulnerabilities of 3D Gaussian Splatting, our work paves the way for more robust and secure 3D reconstructions.
  \keywords{Adversarial Attack \& Defense \and Gaussian Splatting \and Discrete Wavelet Transform}
\end{abstract}

\section{Introduction}
\label{sec:intro}
3D reconstruction, exemplified by Neural Radiance Fields (NeRF) \cite{mildenhall2021nerf} and the more recent 3D Gaussian Splatting (3DGS) \cite{kerbl20233d}, refers to the process of recovering a scene’s spatial geometry and structure from multi-view 2D images. It is a fundamental task in computer vision with wide-ranging applications in robotics \cite{lu2024manigaussian,matsuki2024gaussian,zou20253d,keetha2024splatam,rosinol2023nerf}, autonomous driving \cite{zhou2024drivinggaussian,hess2025splatad,khan2025autosplat,tonderski2024neurad}, and healthcare \cite{cai2024structure,wu2025discretized}.
Especially, 3DGS has recently gained prominence as a dominant approach in 3D vision \cite{bao20253d}. Unlike NeRF, which encodes scenes as continuous volumetric radiance fields parameterized by a multi-layer perceptron, 3DGS explicitly represents the scene using a collection of 3D Gaussian ellipsoids. This explicit modeling confers several advantages over NeRF, including faster rendering speeds, higher visual fidelity, greater interpretability, and improved scalability. These benefits make 3DGS well-suited for deployment on remote servers, where it can generate high-quality scene representations from user-provided images.

\subsection{Vulnerabilities of 3DGS}
While 3DGS offers strong performance, its strengths also reveal critical vulnerabilities. Because it explicitly fits object textures and edges, its rendering quality is highly sensitive to input image quality \cite{lu2025bard}. Moreover, 3DGS does not use a fixed network architecture; instead, its adaptive density control dynamically adjusts the number of Gaussian primitives, creating a flexible, data-dependent parameter space \cite{kerbl20233d}. This explicitness and flexibility make 3DGS particularly vulnerable to adversarial perturbations, which can degrade rendering and increase computational cost.

Although adversarial threats to 3DGS are only beginning to be explored, initial work has emerged. Zeybey \etal \cite{zeybey2024gaussian} introduces a two-stage image-space attack that transfers adversarial noise into 3DGS representations, but it operates solely in 2D and lacks true 3D-level manipulation. In contrast, Poison-Splat \cite{lu2024poison} performs a genuine 3D-level attack by training a surrogate model and applying bi-level optimization, achieving stronger disruption at the cost of significant computational overhead.

\subsection{Why is Defense for 3DGS Underexplored?}
Given the emergence of 3DGS-specific adversarial attacks and the increasing awareness of its inherent vulnerabilities, developing effective defense mechanisms for 3DGS is of growing importance. However, to the best of our knowledge, \textbf{defense strategies against adversarial attacks on 3DGS remain largely underexplored.} This gap stems from several unique obstacles that arise from the intersection of 3DGS’s design and the nature of adversarial threats:

\noindent\textbf{Challenge 1: Incompatibility with Standard Defense Frameworks.}
Most adversarial defenses are tailored for supervised classification models with fixed neural architectures, where bi-level optimization is used (an inner loop crafts adversarial perturbations to maximize loss, while the outer loop minimizes this loss via adversarial training \cite{madry2017towards}). In contrast, 3DGS is a \textit{self-supervised} method \textit{without a fixed network structure}, leading to fundamentally different attack and defense formulations. As a result, standard defense strategies do not apply effectively in this setting.

\noindent\textbf{Challenge 2: Non-Invertible Attack Objectives.}
Even when using a quick proxy workaround, such as a pre-trained 3DGS model, reversing the bi-level optimization objective for defense remains ineffective. This is because adversarial objectives in 3DGS often target \textit{non-invertible} image statistics, such as increasing the total variation of rendered images to inject subtle pixel-level noise. Attempting to reverse such objectives introduces unintended artifacts like image blurring or texture distortion, which ultimately degrade reconstruction quality instead of mitigating the attack.

\noindent\textbf{Challenge 3: Absence of Ground-Truth Clean Data.}
Unlike supervised settings where clean labels are available, 3DGS lacks a clear notion of “ground-truth” clean input. This makes it \textit{impossible to distinguish clean from adversarially perturbed images} during training or inference. As a result, designing defenses that restore adversarial inputs without overcorrecting and harming genuinely clean images becomes highly challenging, especially when performance degradation cannot be tolerated.

\noindent\textbf{Challenge 4: Limited Effectiveness of Traditional 2D Image-Space Defenses.}
Common 2D image-level or frequency-based defense techniques often prove insufficient or harmful in the 3DGS setting. For instance, Gaussian smoothing and bilinear filtering may reduce adversarial perturbations, but they also blur legitimate details critical to accurate scene reconstruction. Similarly, reducing the number of Gaussian primitives can limit adversarial effects but comes at the cost of reconstructive fidelity. Even frequency-domain filtering (e.g., via Fourier transforms) falls short, as it neglects the spatial structure essential to 3D rendering, weakening its ability of meaningful defense for 3DGS.

\begin{figure}
    \centering
    \includegraphics[width=\textwidth]{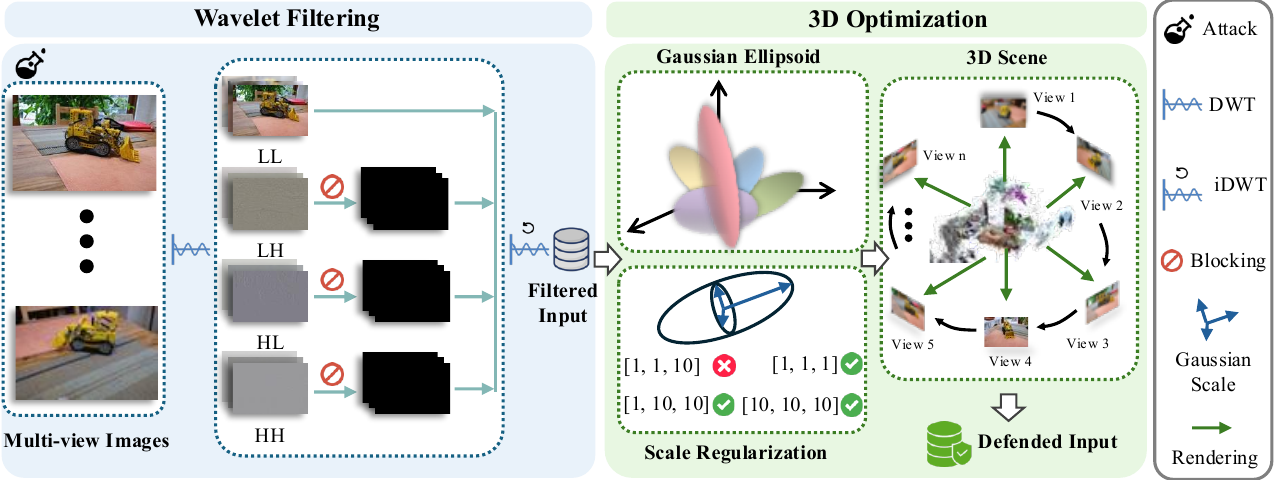}
    \caption{The overview of our proposed method.}
    \label{fig2}
\end{figure}

\subsection{A Spark of Opportunity: Frequency}
In this work, we present the first comprehensive study of defense mechanisms against adversarial attacks on 3DGS. To address the fundamental differences from conventional attack and defense (Challenges 1 and 2), we forgo adversarial training and instead propose a defense strategy that operates directly on the input multi-view images. We begin by analyzing different frequency components of the input images and how adversarial perturbations affect them. Specifically, we employ wavelet transforms to decompose the input into low- and high-frequency components, leveraging their ability to provide both spatial and frequency representations. This enables a spatially-aware, frequency-sensitive consistency analysis across multiple views. Using deep image matching techniques \cite{morelli2024_deep_image_matching}, we evaluate the consistency of these frequency bands and have two important observations, revealing a new opportunity for defense:
\begin{itemize}
    \item \textbf{Observation 1:} The adversarial perturbations on 3DGS inputs often manifest as \textbf{high-frequency noises} (as shown in the right of \cref{fig1}).
    \item \textbf{Observation 2:} The \textbf{low-frequency component retains the main content} of the original scene, as the ratios of energy (a signal for image structure and content) are shown in the bottom left of \cref{fig1}. 
\end{itemize}
Building on this observation, we introduce \textit{DefenseSplat} (as shown in \cref{fig2}), a frequency-aware defense method designed to suppress high-frequency adversarial noises while preserving informative low-frequency content. By doing so, the 3DGS optimization process naturally blurs out inconsistent perturbations during reconstruction, enhancing robustness without requiring access to clean ground-truth data (addressing Challenge 3). Importantly, this approach maintains high reconstruction quality on clean inputs, addressing Challenge 4 by balancing robustness with fidelity, especially in scenarios where significant performance degradation is not acceptable. Our main contribution can be summarized as:
\begin{itemize}
    \item We investigate an important yet previously unexplored problem of defending 3DGS against adversarial attacks \textbf{(in this paper, we consider Poison-Splat as the attack to defend against).} We analyze the significance of this problem and identify the unique challenges it presents.
    \item We introduce a novel frequency-aware defense strategy DefenseSplat for 3DGS by leveraging wavelet analysis to characterize the behavior of adversarial perturbations across frequency bands. To the best of our knowledge, this is the \textit{first} work specifically targeting adversarial defense for 3DGS.
    \item Our comprehensive experiments show that the proposed method enhances the robustness of 3DGS across diverse datasets and a range of attack strengths, without substantially compromising clean-data performance.
\end{itemize}

\section{Related Works}
\label{sec:related}
\subsection{3D Gaussian Splatting.}
3DGS \cite{kerbl20233d} is a recently introduced framework for 3D scene representation, which employs an explicit set of anisotropic 3D Gaussian primitives parameterized by position, scale, orientation, and color to model the underlying geometry and appearance. Unlike implicit methods such as NeRF \cite{mildenhall2021nerf}, 3DGS supports efficient, real-time rendering through a differentiable rasterization process. Owing to its high rendering quality, interpretability, and scalability, 3DGS has been rapidly adopted across a wide spectrum of vision and graphics applications, including dynamic scene reconstruction \cite{wu20244d,yang2024deformable}, open-vocabulary 3D semantic segmentation \cite{wu2024opengaussian,ye2024gaussian,shi2024language,lu2025segment}, and text-to-3D generation. However, as 3DGS is increasingly deployed on cloud-based servers and used in commercial applications, its potential vulnerabilities and susceptibility to adversarial manipulation have become critical concerns.

\noindent\textbf{Adversarial Attack.}
Adversarial attacks are typically formulated as a bi-level optimization problem, where imperceptible perturbations are crafted to maximize the model’s prediction error while remaining constrained in norm. In 2D image classification, such attacks have been extensively studied under white-box and black-box settings. White-box methods \cite{goodfellow2014explaining,madry2017towards,moosavi2016deepfool} utilize gradient access to generate perturbations, whereas black-box attacks \cite{liu2016delving,andriushchenko2020square} rely on transferability or query-based optimization. Due to the inherent vulnerabilities of 3D Gaussian Splatting, several recent works have emerged aiming to exploit these weaknesses through adversarial attacks at the 3D level. Poison-Splat \cite{lu2024poison} is a representative work in this direction, which performs a genuine 3D adversarial attack by pretraining a surrogate 3DGS model and adopting the total variance score of the rendered images as the attack objective. However, adversarial attacks targeting the domain of 3D Gaussian Splatting remain largely underexplored.

\subsection{Defense on Images.}
\looseness -1
The rise of adversarial attacks has spurred advances in defense strategies in images \cite{qiao2025counterfactual}. Among them, adversarial training \cite{madry2017towards, zhang2021causaladv} is the most straightforward and widely used approach. It involves augmenting the training process with adversarial samples, allowing the model to learn parameters that are robust to specific perturbation patterns. However, as mentioned in the challenges discussed in \cref{sec:intro}, adversarial training cannot be directly applied to 3D adversarial attacks. Another line of defense leverages diffusion models to remove artifacts and distortions from images \cite{wu2025difix3d+}. However, such methods typically rely on large-scale pretraining and are primarily designed for scenarios where the degradation in rendering quality arises from sparse-view supervision during 3D reconstruction. As far as we are aware, neither prior work has addressed defense against 3D adversarial attacks, nor are existing defense approaches directly applicable to this task.

\subsection{Frequency Transform.}
The Discrete Fourier Transform (DFT) plays a pivotal role in image processing. Previous works have leveraged DFT on images to enhance synthesis quality \cite{jiang2021focal} and improve image classification accuracy \cite{rao2021global}. More recent studies have extended the application of DFT to the domain of 3D reconstruction. FreGS \cite{zhang2024fregs} leverages additional frequency spectrum supervision to mitigate the issue of over-reconstruction in Gaussian Splatting. 3D-GSW \cite{jang20253d} employs DFT to guide the splitting of 3D Gaussians in high-frequency regions, thereby enhancing reconstruction quality. DFT only captures global frequency information, whereas the Discrete Wavelet Transform (DWT) provides a joint representation of both spatial and frequency domains. DWT can be seamlessly integrated with radiance field representations \cite{xu2023wavenerf,lou2024darenerf,jang2024waterf}, significantly enhancing the quality of 3D reconstruction. Building upon the advantageous properties of DWT and its demonstrated success in 3D reconstruction, we leverage DWT to analyze adversarial attacks on 3DGS from both spatial and frequency perspectives and further design a wavelet-based defense strategy.

\section{Methodology}
\begin{figure}[t]
    \centering
    \includegraphics[width=0.7\columnwidth]{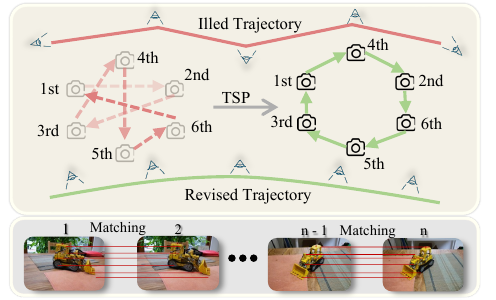}  
    \caption{To accurately measure image matching, we order the camera poses using a Traveling Salesman Problem formulation, which yields an optimized traversal path.}
    \label{fig3}
\end{figure}

\subsection{Preliminaries}
\subsubsection{3D Gaussian Splatting.}
3DGS represents a 3D scene using an explicit set of anisotropic 3D Gaussian primitives. Each primitive is defined as:
\begin{equation}
    G(\mathbf{x};  \boldsymbol{\Sigma}) = \exp\left(-\frac{1}{2} (\mathbf{x})^\top \boldsymbol{\Sigma}^{-1} (\mathbf{x})\right),
\end{equation}
where $\mathbf{x}$ is the center of each primitive. $\boldsymbol{\Sigma}$ is the covariance matrix, which, together with $\mathbf{x}$ determines the spatial distribution of the Gaussian. To enable differentiable optimization, $\boldsymbol{\Sigma}$ is expressed as a combination of a scaling matrix $R$ and a rotation matrix $S$, shown as:
\begin{equation}
    \boldsymbol{\Sigma} = R S S^{\top} R^{\top}.
\end{equation}
Given a specific camera pose $\pi$, each pixel $(x,y)$ color $I_\pi(x,y)$ rendered onto the 2D image is calculated via alpha-blending \cite{kerbl20233d}, taking into account a set of Gaussians ($\mathcal{N}$) that are simultaneously visible and contribute to the pixel through spatial overlap:
\begin{equation}
    I_\pi(x,y) = \sum_{i \in \mathcal{N}} c_i \alpha_i \prod_{j=1}^{i-1} (1 - \alpha_j),
\end{equation}
where $c_i$ is each projected Gaussian given by spherical harmonics, and $\alpha_i$ is the opacity. During the reconstruction training process, the parameters of the 3D Gaussian primitives are optimized by jointly minimizing the $L_1$ loss and the SSIM loss \cite{wang2004image} between the rendered images and the ground-truth images.

\subsubsection{Discrete Wavelet Transform (DWT).} DWT decomposes an image into four frequency subbands: $LL$ (low-low), $LH$ (low-high), $HL$ (high-low), and $HH$ (high-high). The $LL$ subband captures the \textit{low-frequency} components in both horizontal and vertical directions and typically contains the majority of the image’s energy, representing its overall structure and smooth content. This subband can be further recursively decomposed across multiple levels to extract hierarchical low-frequency representations. In contrast, the $LH$, $HL$, and $HH$ subbands capture \textit{high-frequency} information along different orientations: $LH$ highlights horizontal edge-like features by encoding low-frequency horizontal and high-frequency vertical changes; $HL$ emphasizes vertical edges by capturing high-frequency horizontal and low-frequency vertical variations; and $HH$ contains diagonal details with high-frequency variations in both directions. Together, these subbands provide a joint spatial–frequency representation of the image. The DWT \cite{gonzalez2009digital} is defined as:
\begin{align}
W_{\varphi}(j_0, m, n) &= \frac{1}{\sqrt{MN}} \sum_{x=0}^{M-1} \sum_{y=0}^{N-1} I(x, y)\, \varphi_{j_0,m,n}(x, y), \nonumber \\
W_{\psi}^{i}(j, m, n) &= \frac{1}{\sqrt{MN}} \sum_{x=0}^{M-1} \sum_{y=0}^{N-1} I(x, y)\, \psi^{i}_{j,m,n}(x, y),
\end{align}
where $W_{\varphi}$ is the $LL$ subband, and $W_{\psi}^{i}$ represents the $LH$, $HL$, and $HH$ subbands. $j_0$ is the initial approximation scale, $j$ is the scale index for detail subbands. $M$ and $N$ are width and height of the RGB image, while $m$ and $n$ are the spatial location indices in the wavelet coefficient domain. $i=\{H, V, D\}$ denotes the horizontal, vertical, and diagonal coefficients, respectively. $\varphi(x,y)$ is a scaling function and $\psi(x,y)$ is a wavelet function. $I(x,y)$ denotes the pixel intensity of a given position.

\subsection{Frequency-Aware Analysis of Adversarial Perturbations}
\label{sec:freq_analysis}
\subsubsection{Frequency Decomposition.} To design more targeted defense strategies, it is essential to first conduct a comprehensive analysis of the adversarial attacks. Prior works \cite{zhang2024fregs,jang20253d} have demonstrated the power of frequency-domain analysis, establishing it as a powerful tool for enhancing scene reconstruction in 3DGS. For example, it has proven effective in mitigating issues such as over-reconstruction \cite{zhang2024fregs} and supporting applications like watermark embedding \cite{jang20253d}. Building on these successful precedents, we adopt DWT to decompose adversarial images. Owing to its joint representation of both spatial and frequency information, DWT facilitates comprehensive analysis across 2D and 3D domains. Moreover, its frequency decomposition enables a fine-grained dissection of adversarial behavior across different frequency subbands, laying the foundation for targeted defense strategies. Specifically, given an adversarial image $I' \in \mathbb{R}^{3 \times H \times W}$, we apply wavelet decomposition to obtain a set of subbands $\{LL_l, LH_l, HL_l, HH_l\}$, where $l$ denotes the level of DWT. For simplicity and without loss of effectiveness, we set $l = 1$. To constrain each subband within the RGB domain, we perform a wavelet transform on $I'$ channel-wise, normalize each channel to [0, 1], rescale it to [0, 255], and finally fuse the three channels. 

\subsubsection{Vulnerability Analysis.} Next, we use deep image matching \cite{morelli2024_deep_image_matching,morelli2022photogrammetry, ioli2024} to assess the 3D consistency of corresponding subbands between clean and adversarial images, identifying \textit{the most vulnerable regions} where perturbations concentrate. A key challenge before this step is ensuring efficient and accurate multi-view matching. Common strategies include brute-force, retrieval-based, and sequential matching. Brute-force has quadratic complexity, making it inefficient at scale. Retrieval-based methods rely on global features, introducing error without optimal speed. Sequential matching is most efficient but assumes a smooth view sequence, which is often violated in practice due to \textit{randomly ordered} camera poses, leading to failures or reduced accuracy from large viewpoint differences and occlusions.

\begin{table}[t]
\centering
\caption{Quantitative comparison on {Mip-NeRF 360} and {Tanks-and-Temples datasets}. Colors indicate the \shadeText{red}{best} and \shadeText{orange}{second best} results. \#G/M indicates the number of Gaussians (Million).}
\resizebox{\linewidth}{!}{
\begin{tabular}{l|ccc|cccc}
\toprule
Method & Training Time $\downarrow$  & \#G/M $\downarrow$ & GPU Mem (MB) $\downarrow$ & PSNR $\uparrow$ & SSIM $\uparrow$ & LPIPS $\downarrow$ & FPS $\uparrow$ \\
\midrule
\multicolumn{8}{c}{\cellcolor{gray!20}\textit{Mip-NeRF 360}} \\
\midrule
3DGS         & 1:01:01 & 5.91 & 20965 & 25.18 & 0.6356 & 0.4481 & 70 \\
Compact GS   & 1:22:26 & 3.00 & 21072 & 24.95 & 0.6374 & 0.4590 & 128 \\
Difix        & 0:45:44 & 3.65 & 15894 & 24.52 & 0.7326 & 0.3575 & 137 \\
\midrule
Ours         & \cellcolor[HTML]{F9CB9C}0:34:26 & \cellcolor[HTML]{F9CB9C}2.24 & \cellcolor[HTML]{F9CB9C}12567 & \cellcolor[HTML]{F9CB9C}27.32 & \cellcolor[HTML]{F9CB9C}0.7968 & \cellcolor[HTML]{F9CB9C}0.3469 & \cellcolor[HTML]{F9CB9C}243 \\
Ours+ReLU    & \cellcolor[HTML]{EA9999}0:33:49 & \cellcolor[HTML]{EA9999}2.16 & \cellcolor[HTML]{EA9999}11400 & \cellcolor[HTML]{EA9999}27.47 & \cellcolor[HTML]{EA9999}0.8025 & \cellcolor[HTML]{EA9999}0.3417 & \cellcolor[HTML]{EA9999}258 \\
\midrule
\multicolumn{8}{c}{\cellcolor{gray!20}\textit{Tanks-and-Temples}} \\
\midrule
3DGS         & 0:31:22 & 2.76 & 10089 & 24.88 & 0.6425 & 0.4190 & 165 \\
Compact GS   & 0:43:29 & 1.73 & 10930 & 24.37 & 0.6540 & 0.4278 & 244 \\
Difix        & 0:24:12 & 1.71 & 7749  & 24.75 & 0.8092 & \cellcolor[HTML]{EA9999}0.3031 & 290 \\
\midrule
Ours         & \cellcolor[HTML]{EA9999}0:18:56 & \cellcolor[HTML]{F9CB9C}1.06 & \cellcolor[HTML]{F9CB9C}6296  & \cellcolor[HTML]{F9CB9C}26.29 & \cellcolor[HTML]{F9CB9C}0.8281 & 0.3199 & \cellcolor[HTML]{F9CB9C}468 \\
Ours+ReLU    & \cellcolor[HTML]{F9CB9C}0:19:19 & \cellcolor[HTML]{EA9999}1.01 & \cellcolor[HTML]{EA9999}5858  & \cellcolor[HTML]{EA9999}26.42 & \cellcolor[HTML]{EA9999}0.8373 & \cellcolor[HTML]{F9CB9C}0.3134 & \cellcolor[HTML]{EA9999}498 \\
\bottomrule
\end{tabular}
}
\label{tab1}
\end{table}

To address this issue, we formulate it as a Traveling Salesman Problem (TSP). We first convert each camera pose into the form of the Special Euclidean group in 3D, and then compute the pose loss between every pair of cameras, defined as:
\begin{equation}
    \mathcal{L}_{\text{pose}} = w_g \, \mathcal{L}_{\text{geo}} + w_t \, \mathcal{L}_{\text{trans}}, \label{eq:pose_loss}
\end{equation}
where $\mathcal{L}_{\text{geo}}$ is the geodesic loss, $\mathcal{L}_{\text{trans}}$ is the translation loss, $w_g$ and $w_t$ are their corresponding weights. Specifically, these two losses are defined as:
\begin{gather}
\mathcal{L}_{\text{geo}} = \arccos\left( \frac{\mathrm{trace}\!\left( R_{\text{pred}}^\top R_{\text{gt}} \right) - 1}{2} \right),  \\
\mathcal{L}_{\text{trans}} = \left\| t_{\text{pred}} - t_{\text{gt}} \right\|_2^2,
\end{gather}
where $R_{(\cdot)}$ and $t_{(\cdot)}$ represent the rotation matrix and the translation vector, respectively. The subscripts indicate prediction and ground truth, respectively. We treat each camera as a city and the pose loss between cameras as the distance between cities in TSP. By solving the resulting instance of the TSP using the Lin–Kernighan \cite{lin1973effective} algorithm, we obtain a smooth and optimized camera trajectory (as shown in \cref{fig3}). Based on the optimized trajectory, we adopt the state-of-the-art SuperPoint \cite{detone2018superpoint} as the feature extractor and LightGlue \cite{lindenberger2023lightglue} as the matcher to perform pairwise matching across all wavelet subbands of clean and adversarial images. Each image is only matched with subsequent images along the trajectory. For each image pair, the \textit{matching rate} is defined as the ratio of the number of matched keypoints to the number of extracted keypoints, indicating the \textit{consistency of the image across multiple viewpoints}. The overall multi-view matching rate is computed by averaging the pairwise matching rates. Since the $HH$ subband contains the least energy (only $<0.08\%$ in almost all datasets, with details in Appendix B) and diagonal details are relatively rare in natural images, we omit $HH$ and use $LH$ and $HL$ to represent the high-frequency information. As shown in \cref{fig1}, both the bar charts and subband visualizations indicate that \textbf{the vulnerability is concentrated in the high-frequency regions}. The consistency degradation in high-frequency subbands is significantly more pronounced than in low-frequency ones, suggesting that \textbf{3D adversarial attacks primarily target high-frequency components by injecting random noise, while introducing relatively smooth and consistent pseudo-textures in the low-frequency regions.}

\subsection{Defense via Frequency-Based Filtering}
\label{sec:freq_filter}
Based on the above analysis and insights, we propose a simple yet effective defense strategy, DefenseSplat. Given adversarial images, we first apply DWT to obtain four subbands. The high-frequency subbands (\textit{i.e.}, $LH$, $HL$, and $HH$) are then set to \textit{zero}, and the inverse wavelet transform is performed using the LL subband together with the zeroed high-frequency subbands to reconstruct the filtered image in the spatial domain. This process can be defined as:
\begin{equation}
    I_f=\texttt{iDWT}(\texttt{DWT}(I')_{LL},0,0,0),
\end{equation}
where $I_f$ represents the filtered images, $\texttt{iDWT}(\cdot)$ is the inverse Discrete Wavelet Transform, $I'$ denotes the adversarial images and $\texttt{DWT}(I')_{LL}$ is the $LL$ subband extracted from $I'$. Merely relying on the $I_f$ component is far from sufficient, as it still contains artificial textures with relatively low consistency in the low-frequency domain. To mitigate this problem, we exploit an intrinsic property of 3DGS: when fitting regions with low cross-view consistency, the Gaussians tend to average the colors across views to minimize the $L_1$ loss, thereby producing a blurred reconstruction over these regions, which naturally further suppresses adversarial artifacts in $I_f$ that lack consistent support across views.

Another common observation is: For certain artificially injected textures that exhibit high multi-view consistency, the reconstruction process tends to generate \textit{elongated} Gaussians to fit these patterns, which increases the number of Gaussians and memory consumption, thereby undermining the effectiveness of the defense. To tackle this issue, we design a ReLU-based scale regularization loss on the normalized variance of the scales along the three principal axes of each Gaussian, defined as follows:
\begin{equation}
\label{scale loss}
    \mathcal{L}_{scale}=\text{ReLU}(\nu-\tau),
\end{equation}
where $\nu$ denotes the normalized variance, $\tau$ is a predefined threshold that limits the range of scale regularization. For Gaussians whose normalized variance falls below $\tau$, the loss does not propagate gradients. This loss \textit{constrains elongated Gaussians without affecting the other two types of Gaussians}: small spherical Gaussians and large flat Gaussians (because their normalized variance of scales does not exceed that of elongated Gaussians). Both of them are crucial: the former contributes to reconstructing fine details and alleviates over-reconstruction, while the latter enhances the reconstruction quality of large smooth regions and mitigates under-reconstruction. Therefore, this loss can effectively suppress adversarial textures while preserving the reconstruction of genuine scene details. Finally, after training, the 3D Gaussians are rendered on all camera views to generate the final defended images for downstream tasks.

\subsection{Fine-Grained Cross-View Consistency Filtering (Optional)}
\label{sec:fine_grained}

The coarse filter in \cref{sec:freq_filter} removes all high-frequency coefficients, effectively suppressing adversarial perturbations but also discarding genuine scene details and causing over-smoothed reconstructions on structured regions. To avoid this all-or-nothing suppression, we introduce a fine-grained cross-view-consistency filter. We first order all camera poses using the TSP tour from \cref{sec:freq_analysis}. Each adversarial image is then compared with its left neighbour along this pose-ordered cycle, with the first pose matched back to the last. For each high-frequency wavelet subband, we retain coefficients around locations that match the corresponding subband in the neighbouring view, treating them as multi-view consistent scene details, and suppress the unmatched coefficients as view-inconsistent perturbations. The defended image is finally reconstructed by keeping the low-frequency subband unchanged and inverting the wavelet transform with only the matched high-frequency coefficients. This preserves authentic details erased by the coarse filter while removing adversarial components that do not recur across nearby views.

\section{Experiments}
\begin{table}[t]
\centering
\caption{Quantitative comparison under different attack levels on Tanks-and-Temples.}
\resizebox{\linewidth}{!}{
\begin{tabular}{cc|ccc|ccccc}
\toprule
$\epsilon$ & Method & Training Time ↓ & \#G/M ↓ & GPU Mem (MB) ↓ & PSNR ↑ & SSIM ↑ & LPIPS ↓ & FPS ↑ \\
\midrule
\multirow{5}{*}{16/255} 
  & 3DGS       & 0:27:38 & 2.17 & 8558  & 24.03 & 0.662 & 0.375 & 215 \\
  & Compact GS      & 0:36:58 & 1.47 & 9466  & 23.59 & 0.669 & 0.386 & 306 \\
  & Difix           & 0:21:39 & 1.48 & 7238  & 23.59 & 0.787 & 0.298 & 295 \\
  & Ours            & \cellcolor[HTML]{EA9999}0:17:44 & \cellcolor[HTML]{F9CB9C}0.99 & \cellcolor[HTML]{F9CB9C}5642  & \cellcolor[HTML]{F9CB9C}24.89 & \cellcolor[HTML]{F9CB9C}0.815 & \cellcolor[HTML]{F9CB9C}0.294 & \cellcolor[HTML]{F9CB9C}449 \\
  & Ours + ReLU     & \cellcolor[HTML]{F9CB9C}0:18:17 & \cellcolor[HTML]{EA9999}0.96 & \cellcolor[HTML]{EA9999}5418  & \cellcolor[HTML]{EA9999}25.03 & \cellcolor[HTML]{EA9999}0.824 & \cellcolor[HTML]{EA9999}0.291 & \cellcolor[HTML]{EA9999}473 \\
\midrule
\multirow{5}{*}{32/255} & 3DGS       & 0:36:14 & 3.24 & 11541 & 20.36 & 0.449 & 0.485 & 140 \\
   & Compact GS      & 0:50:13 & 2.20 & 12997 & 20.26 & 0.458 & 0.491 & 199 \\
   & Difix           & 0:23:16 & 1.69 & 7417  & 22.22 & \cellcolor[HTML]{EA9999}0.707 & \cellcolor[HTML]{EA9999}0.348 & 263 \\
   & Ours            & \cellcolor[HTML]{F9CB9C}0:20:02 & \cellcolor[HTML]{F9CB9C}1.25 & \cellcolor[HTML]{F9CB9C}6182  & \cellcolor[HTML]{F9CB9C}23.47 & 0.690 & 0.377 & 362 \\
   & Ours + ReLU     & \cellcolor[HTML]{EA9999}0:19:43 & \cellcolor[HTML]{EA9999}1.19 & \cellcolor[HTML]{EA9999}5806  & \cellcolor[HTML]{EA9999}23.61 & \cellcolor[HTML]{F9CB9C}0.703 & \cellcolor[HTML]{F9CB9C}0.374 & \cellcolor[HTML]{EA9999}388 \\
\midrule
\multirow{5}{*}{64/255} & 3DGS       & 0:43:00 & 4.07 & 12418 & 15.55 & 0.272 & 0.586 & 119 \\
   & Compact GS      & 0:59:48 & 2.69 & 15807 & 15.65 & 0.285 & 0.591 & 178 \\
   & Difix           & 0:25:46 & 1.99 & 8067  & 19.18 & \cellcolor[HTML]{EA9999}0.504 & \cellcolor[HTML]{EA9999}0.458 & 243 \\
   & Ours            & \cellcolor[HTML]{F9CB9C}0:23:42 & \cellcolor[HTML]{F9CB9C}1.79 & \cellcolor[HTML]{F9CB9C}7651  & \cellcolor[HTML]{F9CB9C}19.27 & 0.443 & 0.509 & 281 \\
   & Ours + ReLU     & \cellcolor[HTML]{EA9999}0:23:16 & \cellcolor[HTML]{EA9999}1.71 & \cellcolor[HTML]{EA9999}6958  & \cellcolor[HTML]{EA9999}19.38 & \cellcolor[HTML]{F9CB9C}\cellcolor[HTML]{F9CB9C}0.454 & \cellcolor[HTML]{F9CB9C}0.505 & \cellcolor[HTML]{EA9999}301 \\
\bottomrule
\end{tabular}
}
\label{tab2}
\end{table}
\begin{table}[t]
\centering
\caption{Quantitative results of our method on clean inputs.}
\resizebox{\linewidth}{!}{
\begin{tabular}{l|ccc|cccc}
\toprule
Method & Training Time $\downarrow$ & \#G/M $\downarrow$ & GPU Mem (MB) $\downarrow$ & PSNR $\uparrow$ & SSIM $\uparrow$ & LPIPS $\downarrow$ & FPS $\uparrow$ \\
\midrule
\multicolumn{8}{c}{\cellcolor{gray!20}\textit{Mip-NeRF 360}} \\
\midrule
Clean               & 0:38:20 & 2.62 & 20965 & 29.06 & 0.868 & 0.248  & 223 \\
Clean + Ours        & 0:32:11 & 1.83 & 11012 & 28.58 & 0.844 & 0.280  & 314 \\
Clean + Ours + ReLU & 0:32:23 & 1.79 & 10945 & 28.79 & 0.847 & 0.271  & 321 \\
\midrule
\multicolumn{8}{c}{\cellcolor{gray!20}\textit{Tanks-and-Temples}} \\
\midrule
Clean               & 0:22:48 & 1.56 & 7328  & 28.06 & 0.898 & 0.212  & 327 \\
Clean + Ours        & 0:18:24 & 0.97 & 6161  & 27.18 & 0.868 & 0.254  & 517 \\
Clean + Ours + ReLU & 0:18:34 & 0.94 & 5857  & 27.33 & 0.875 & 0.245  & 526 \\
\bottomrule
\end{tabular}
}
\label{tab3}
\end{table}

In this section, we conduct extensive experiments on multiple datasets. Specifically, we address the following research questions based on the experimental results. \textbf{RQ1:} How does DefenseSplat perform compared to other baselines? Is it an effective defense strategy? \textbf{RQ2:} Can DefenseSplat consistently demonstrate effective defense capability under varying levels of attack intensity? 
\textbf{RQ3:} Given the diversity of user-submitted inputs, can DefenseSplat maintain a balance between robustness and performance on clean data? \textbf{RQ4:} Can the Gaussian scale loss (\cref{scale loss}) further enhance the overall performance of DefenseSplat (ablation study)? \textbf{RQ5:} Can our method remain resilient to stronger attacks, such as the recently proposed Adaptive Attack\cite{li2025remedygs}? \textbf{RQ6:} Can the fine-grained filter improve reconstruction performance?

\begin{figure}
    \centering
    \includegraphics[width=0.9\textwidth]{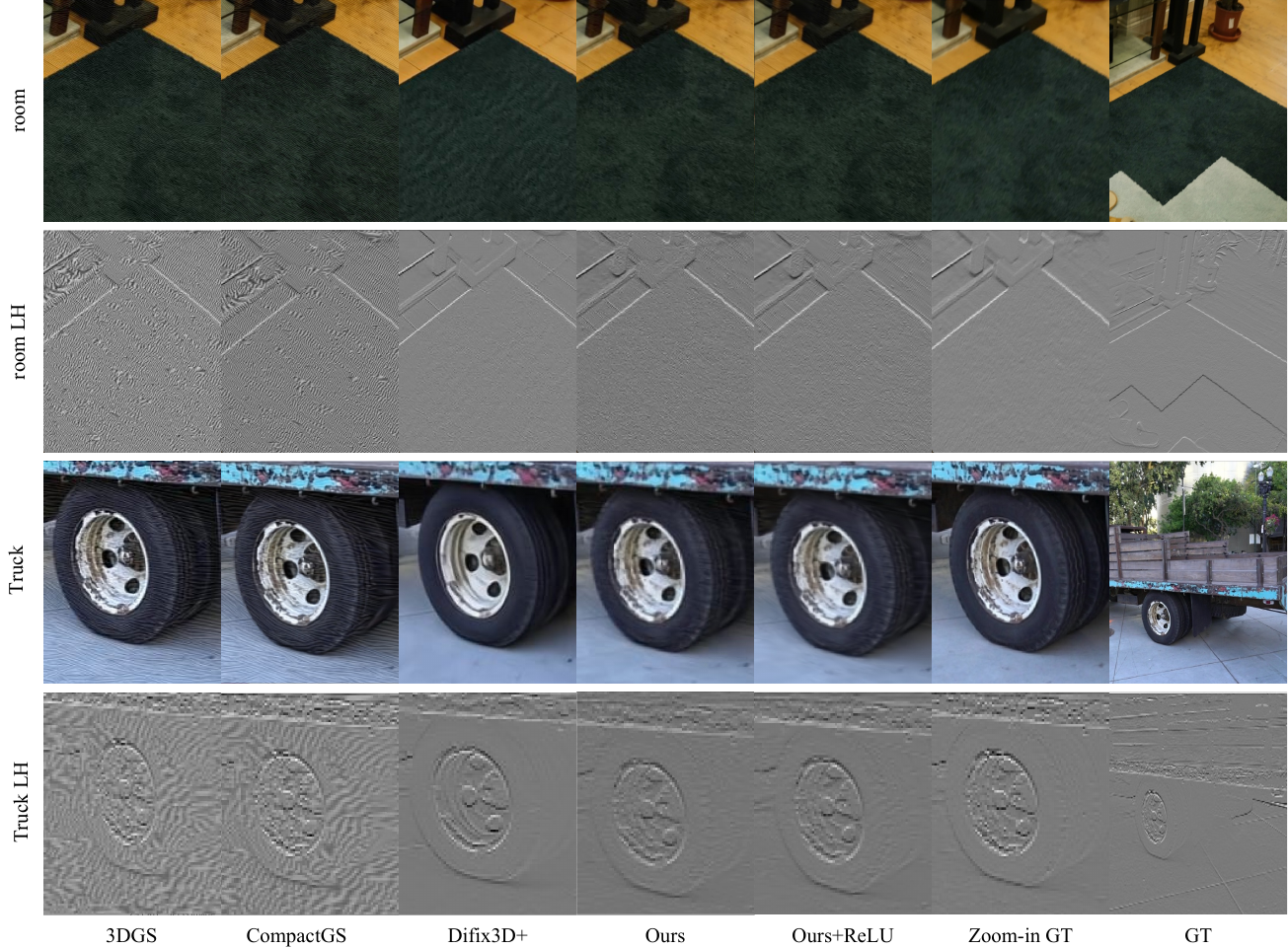} 
    \caption{Comparison of reconstruction quality of all methods and ground truth (GT) on Mip-NeRF 360 and Tanks-and-Temples datasets.}
    \label{fig4}
\end{figure}

\subsection{Experiment Settings}
\textbf{Datasets.} We use Mip-NeRF 360 \cite{barron2022mip}, Tanks-and-Temples \cite{Knapitsch2017} and LLFF \cite{mildenhall2019local}, which are commonly used benchmarks in the field of 3D reconstruction. For the Tanks and Temples dataset, we exclude scenes containing dynamic objects and retain 8 static scenes for evaluation. For Mip-NerF 360 and LLFF datasets, we include all available scenes.

\noindent\textbf{Baselines}. Due to the absence of prior research dedicated to the defense of 3D adversarial attacks, no existing method serves as a directly comparable baseline. Therefore, in addition to the original \textbf{3DGS} as a baseline, we select two other methods whose research directions are most closely related to defense as comparative baselines. The first one is \textbf{Difix3D+} \cite{wu2025difix3d+}, which leverages diffusion models to enhance the quality of reconstructed images while mitigating 3D inconsistency. Moreover, it shares a similar setting with ours, as both methods operate directly on the input images. The second is \textbf{CompactGS} \cite{lee2024compact}, which achieves reduced storage and faster rendering speed while maintaining high-quality reconstruction by effectively removing redundant Gaussians that do not significantly contribute to the overall performance. This objective closely aligns with the essence of defense.

\noindent\textbf{Implementation Details.} 
Our method and all baselines are trained on the same server equipped with four A6000 GPUs. To ensure fairness, all methods are trained for 30,000 iterations. The remaining hyperparameters of the baselines are kept unchanged from their original settings. Unless otherwise specified, we use adversarial images with an attack strength of $\epsilon=16/255$ as the training input. The weight for the scale loss $\lambda_{scale}$ is set to $1 \times 10^5$.

\noindent\textbf{Evaluation Metrics} We evaluate the effectiveness of our defense strategy along two dimensions: \textbf{robustness} and \textbf{reconstruction fidelity}. For robustness, we evaluate training time, the number of Gaussians, peak GPU memory usage, and rendering speed, which together reflect the server’s capacity to sustain training under the load imposed by the input images. For reconstruction fidelity, we adopt widely used metrics including peak
signal-to-noise ratio (PSNR), structural similarity (SSIM), and learned perceptual image patch similarity (LPIPS).

\subsection{Experiment Results}
\noindent\textbf{Defense Effectiveness of DefenseSplat (RQ1).}
We conduct a comprehensive comparison between our method and the baselines. As shown in \cref{tab1}, our method \textbf{achieves the best performance in both robustness and rendering quality} on the Mip-NeRF 360 and the Tanks-and-Temples datasets with the sole exception of the LPIPS on the Tanks-and-Temples dataset, where our score is only slightly higher than the best-performing baseline. However, this does not indicate a performance drawback: our method still achieves the lowest peak GPU memory usage—an essential metric, as exceeding hardware limits can lead to denial-of-service issues and prevent rendering. \cref{fig4} presents visualizations of the rendered images. Since the perturbations introduced by the 3D adversarial attacks are nearly imperceptible to humans, we zoom in to better highlight the details. As observed in both the room and truck scenes, the rendered images from the original 3DGS contain numerous adversarial artifacts, which are more clearly revealed in the corresponding $LH$ subband. Although CompactGS reduces the number of Gaussians, it still fails to effectively constrain the Gaussians from overfitting to adversarial perturbations. The rendered images produced by Difix3D+ appear smoother; however, they tend to introduce new spurious textures or remove authentic ones. In the room scene, wavy artifacts emerge on the carpet, while in the Truck scene, genuine details such as the rust on the hub and the tire tread patterns are mistakenly eliminated. Our method effectively removes adversarial artifacts while preserving the original textures to the greatest extent, resulting in reconstructions that are closest to the ground truth.

\looseness -1
\noindent\textbf{Defense Effectiveness Under Various Attack Strengths (RQ2).} As shown in \cref{tab2}, our method consistently \textbf{shows effective defense under varying attack strengths}, due to the fact that the attack primarily targets high-frequency information, which our method blocks. Even when the attack strength increases and low-frequency components are possibly affected, our defense still remains effective, since our method consistently filters out high-frequency content.

\looseness -1
\noindent\textbf{Robustness and Fidelity Trade-Off (RQ3).} As shown in \cref{tab3}, even when the input images are clean, our method \textbf{does not exhibit any significant performance degradation}. The PSNR and SSIM drop by only 1.75\% and 2.5\%, respectively, and LPIPS increases by just 11\%. Although our method removes high-frequency information, the $LL$ subband retains over 95\% of the total energy across all subbands. As a result, even for clean images, the majority of information is preserved, leading to no significant degradation in reconstruction quality. 

\noindent\textbf{Effectiveness of Scale Loss (RQ4).} As evidenced by \cref{tab1} and \cref{tab3}, introducing the scale loss leads to a noticeable improvement in our method’s performance, yielding the best results across most evaluation metrics. We provide additional experimental results in Appendix B.

\begin{table}[t]
\centering
\caption{Results of Adaptive Attack on two scenes of NeRF Synthetic Dataset (TT=Training Time, $\#G$=Number of Gaussians, GPU=GPU Memory)}
{\fontsize{8pt}{9.6pt}\selectfont
\begin{tabular}{ccccccccc}
\hline
Scene                  & \multicolumn{1}{c|}{Settings} & TT      & \#G    & GPU  & PSNR  & SSIM   & LPIPS  & FPS  \\ \hline
\multicolumn{9}{c}{\cellcolor{gray!20}\textit{$\beta=1$}}                                                                                        \\ \hline
\multirow{2}{*}{chair} & \multicolumn{1}{c|}{Attack}   & 0:16:00 & 889700 & 6052 & 30.14 & 0.3483 & 0.3971 & 436  \\
                       & \multicolumn{1}{c|}{Ours}     & 0:08:06 & 200798 & 2338 & 30.31 & 0.4123 & 0.3241 & 1653 \\ \hline
\multirow{2}{*}{ficus} & \multicolumn{1}{c|}{Attack}   & 0:09:17 & 280991 & 5982 & 30.44 & 0.3015 & 0.4365 & 1276 \\
                       & \multicolumn{1}{c|}{Ours}     & 0:06:07 & 121204 & 2118 & 31.68 & 0.3405 & 0.3589 & 2381 \\ \hline
\multicolumn{9}{c}{\cellcolor{gray!20}\textit{$\beta=5$}}                                                                                        \\ \hline
\multirow{2}{*}{chair} & \multicolumn{1}{c|}{Attack}   & 0:16:43 & 897341 & 6186 & 30.04 & 0.3428 & 0.4051 & 437  \\
                       & \multicolumn{1}{c|}{Ours}     & 0:08:10 & 202392 & 2182 & 30.32 & 0.4085 & 0.3337 & 1654 \\ \hline
\multirow{2}{*}{ficus} & \multicolumn{1}{c|}{Attack}   & 0:09:17 & 280642 & 6194 & 30.41 & 0.3012 & 0.4396 & 1288 \\
                       & \multicolumn{1}{c|}{Ours}     & 0:06:04 & 121065 & 2120 & 31.69 & 0.3438 & 0.3627 & 2358 \\ \hline
\multicolumn{9}{c}{\cellcolor{gray!20}\textit{Clean}}                                                                                         \\ \hline
\multicolumn{2}{c|}{chair}                             & 0:09:17 & 439996 & 2846 & 39.08 & 0.9948 & 0.0083 & 892  \\
\multicolumn{2}{c|}{ficus}                             & 0:06:07 & 170861 & 1818 & 38.00 & 0.9954 & 0.0080 & 1930 \\ \hline
\end{tabular}}
\label{tab_attack}
\end{table}
\noindent\textbf{Defense Effectiveness against Adaptive Attack (RQ5).} Adaptive Attack is constructed by adding a weighted term, the mutual information between the clean and the poisoned images, to the PoisonSplat objective. This modification makes the poisoned images more imperceptible and harder to defend against. The formulation is as follows:
\begin{equation}
    \max_{\mathcal{V}_{\mathrm{poi}}} \mathcal{L} = \max_{\mathcal{V}_{\mathrm{poi}}} \{ \mathcal{S}_{TV}(\mathcal{V}_{\mathrm{poi}}) + \beta \cdot I(\mathcal{V}_{\mathrm{cln}}; \mathcal{V}_{\mathrm{poi}}) \},
\end{equation}
\looseness -1
where $\mathcal{S}_{TV}$ denotes the TV score, $I(\mathcal{V}_{\mathrm{cln}}; \mathcal{V}_{\mathrm{poi}})$ denotes the mutual information between the clean images and poisoned images, $\beta$ is the weight of mutual information. Since Adaptive Attack does not report the exact number of attack iterations, a fair comparison is not possible; therefore, it is not the primary attack method we focus on. We conduct defense evaluations on two scenes from the NeRF Synthetic dataset. As shown in \cref{tab_attack}, our method provides effective defense against Adaptive Attack across different strength levels ($\beta$). In our experiments, we adopt an extremely large number of attack steps (36,000), which leads to relatively worse post-attack SSIM and LPIPS values. Under this severe setting, our method still provides effective defense and does not require access to the clean images, which is more reasonable.

\begin{table}[t]
\centering
\caption{Quantitative results comparison of basic filter and fine-grained filter.}
\resizebox{\linewidth}{!}{
\begin{tabular}{l|ccc|cccc}
\toprule
Method & Training Time $\downarrow$ & \#G/M $\downarrow$ & GPU Mem (MB) $\downarrow$ & PSNR $\uparrow$ & SSIM $\uparrow$ & LPIPS $\downarrow$ & FPS $\uparrow$ \\
\midrule
basic               & 0:33:49 & 2.16 & 11400 & 27.47 & 0.8025 & 0.3417  & 258 \\
fine-grained        & 0:34:15 & 2.19 & 11373 & 27.49 & 0.8027 & 0.3388  & 265 \\
\bottomrule
\end{tabular}
}
\label{tab_fine_grained}
\end{table}
\noindent\textbf{Effectiveness of the fine-grained filter (RQ6).} We evaluate both filters on the Mip-NeRF 360 dataset and report the average results. As shown in  \cref{tab_fine_grained}, the fine-grained filter slightly improves reconstruction quality, validating our previous analysis and the effectiveness of the proposed method. However, since the improvement brought by the fine-grained filter is limited and its data preprocessing is more complex, this component can be optional when the overall efficiency is preferred.

\section{Discussion and Conclusion}
In this work, we propose a novel and effective defense strategy DefenseSplat against 3DGS adversarial attacks. By applying filtering on high-frequency wavelet subbands, our method achieves both robust 3D reconstruction training on the server side and faithful image rendering quality. Since only one main attack and its variant on 3DGS have been proposed so far, our evaluation is necessarily limited to this specific method. However, given our observation that adversarial attacks primarily target vulnerable high-frequency components in the input, we believe our defense strategy is broadly generalizable. We remain confident and look forward to testing its effectiveness against future, unseen attacks. It is worth noting that our defense strategy can also be flexibly employed as a plug-and-play module, allowing seamless integration with existing methods to achieve better defense performance.

%
%
\bibliographystyle{splncs04}
\bibliography{main}

@String(TOG   = {ACM Trans. Graph.})

@String(TOG   = {ACM TOG})

@inproceedings{matsuki2024gaussian,
  title={Gaussian splatting slam},
  author={Matsuki, Hidenobu and Murai, Riku and Kelly, Paul HJ and Davison, Andrew J},
  booktitle={Proceedings of the IEEE/CVF Conference on Computer Vision and Pattern Recognition},
  pages={18039--18048},
  year={2024}
}

@inproceedings{lu2024manigaussian,
  title={Manigaussian: Dynamic gaussian splatting for multi-task robotic manipulation},
  author={Lu, Guanxing and Zhang, Shiyi and Wang, Ziwei and Liu, Changliu and Lu, Jiwen and Tang, Yansong},
  booktitle={European Conference on Computer Vision},
  pages={349--366},
  year={2024},
  organization={Springer}
}

@inproceedings{zou20253d,
  title={3D-SPATIAL MULTIMODAL MEMORY},
  author={Zou, Xueyan and Song, Yuchen and Qiu, Ri-Zhao and Peng, Xuanbin and Ye, Jianglong and Liu, Sifei and Wang, Xiaolong},
  booktitle={The Thirteenth International Conference on Learning Representations},
  year={2025}
}

@inproceedings{keetha2024splatam,
  title={Splatam: Splat track \& map 3d gaussians for dense rgb-d slam},
  author={Keetha, Nikhil and Karhade, Jay and Jatavallabhula, Krishna Murthy and Yang, Gengshan and Scherer, Sebastian and Ramanan, Deva and Luiten, Jonathon},
  booktitle={Proceedings of the IEEE/CVF Conference on Computer Vision and Pattern Recognition},
  pages={21357--21366},
  year={2024}
}

@inproceedings{rosinol2023nerf,
  title={Nerf-slam: Real-time dense monocular slam with neural radiance fields},
  author={Rosinol, Antoni and Leonard, John J and Carlone, Luca},
  booktitle={2023 IEEE/RSJ International Conference on Intelligent Robots and Systems (IROS)},
  pages={3437--3444},
  year={2023},
  organization={IEEE}
}

@inproceedings{zhou2024drivinggaussian,
  title={Drivinggaussian: Composite gaussian splatting for surrounding dynamic autonomous driving scenes},
  author={Zhou, Xiaoyu and Lin, Zhiwei and Shan, Xiaojun and Wang, Yongtao and Sun, Deqing and Yang, Ming-Hsuan},
  booktitle={Proceedings of the IEEE/CVF conference on computer vision and pattern recognition},
  pages={21634--21643},
  year={2024}
}

@inproceedings{hess2025splatad,
  title={Splatad: Real-time lidar and camera rendering with 3d gaussian splatting for autonomous driving},
  author={Hess, Georg and Lindstr{\"o}m, Carl and Fatemi, Maryam and Petersson, Christoffer and Svensson, Lennart},
  booktitle={Proceedings of the Computer Vision and Pattern Recognition Conference},
  pages={11982--11992},
  year={2025}
}

@inproceedings{khan2025autosplat,
  title={Autosplat: Constrained gaussian splatting for autonomous driving scene reconstruction},
  author={Khan, Mustafa and Fazlali, Hamidreza and Sharma, Dhruv and Cao, Tongtong and Bai, Dongfeng and Ren, Yuan and Liu, Bingbing},
  booktitle={2025 IEEE International Conference on Robotics and Automation (ICRA)},
  pages={8315--8321},
  year={2025},
  organization={IEEE}
}

@inproceedings{tonderski2024neurad,
  title={Neurad: Neural rendering for autonomous driving},
  author={Tonderski, Adam and Lindstr{\"o}m, Carl and Hess, Georg and Ljungbergh, William and Svensson, Lennart and Petersson, Christoffer},
  booktitle={Proceedings of the IEEE/CVF Conference on Computer Vision and Pattern Recognition},
  pages={14895--14904},
  year={2024}
}

@inproceedings{wu2025discretized,
  title={Discretized Gaussian Representation for Tomographic Reconstruction},
  author={Wu, Shaokai and Lu, Yuxiang and Guo, Yapan and Ji, Wei and Huang, Suizhi and Yang, Fengyu and Sirejiding, Shalayiding and He, Qichen and Tong, Jing and Ji, Yanbiao and others},
  booktitle={Proceedings of the IEEE/CVF International Conference on Computer Vision},
  pages={25073--25082},
  year={2025}
}

@inproceedings{cai2024structure,
  title={Structure-aware sparse-view x-ray 3d reconstruction},
  author={Cai, Yuanhao and Wang, Jiahao and Yuille, Alan and Zhou, Zongwei and Wang, Angtian},
  booktitle={Proceedings of the IEEE/CVF conference on computer vision and pattern recognition},
  pages={11174--11183},
  year={2024}
}

@article{mildenhall2021nerf,
  title={Nerf: Representing scenes as neural radiance fields for view synthesis},
  author={Mildenhall, Ben and Srinivasan, Pratul P and Tancik, Matthew and Barron, Jonathan T and Ramamoorthi, Ravi and Ng, Ren},
  journal={Communications of the ACM},
  volume={65},
  number={1},
  pages={99--106},
  year={2021},
  publisher={ACM New York, NY, USA}
}

@article{kerbl20233d,
  title={3D Gaussian splatting for real-time radiance field rendering.},
  author={Kerbl, Bernhard and Kopanas, Georgios and Leimk{\"u}hler, Thomas and Drettakis, George},
  journal={ACM Trans. Graph.},
  volume={42},
  number={4},
  pages={139--1},
  year={2023}
}

@article{bao20253d,
  title={3d gaussian splatting: Survey, technologies, challenges, and opportunities},
  author={Bao, Yanqi and Ding, Tianyu and Huo, Jing and Liu, Yaoli and Li, Yuxin and Li, Wenbin and Gao, Yang and Luo, Jiebo},
  journal={IEEE Transactions on Circuits and Systems for Video Technology},
  year={2025},
  publisher={IEEE}
}

@article{lu2024poison,
  title={Poison-splat: Computation cost attack on 3d gaussian splatting},
  author={Lu, Jiahao and Zhang, Yifan and Shen, Qiuhong and Wang, Xinchao and Yan, Shuicheng},
  journal={arXiv preprint arXiv:2410.08190},
  year={2024}
}

@inproceedings{lu2025bard,
  title={Bard-gs: Blur-aware reconstruction of dynamic scenes via gaussian splatting},
  author={Lu, Yiren and Zhou, Yunlai and Liu, Disheng and Liang, Tuo and Yin, Yu},
  booktitle={Proceedings of the Computer Vision and Pattern Recognition Conference},
  pages={16532--16542},
  year={2025}
}

@article{zeybey2024gaussian,
  title={Gaussian Splatting Under Attack: Investigating Adversarial Noise in 3D Objects},
  author={Zeybey, Abdurrahman and Ergezer, Mehmet and Nguyen, Tommy},
  journal={arXiv preprint arXiv:2412.02803},
  year={2024}
}

@article{madry2017towards,
  title={Towards deep learning models resistant to adversarial attacks},
  author={Madry, Aleksander and Makelov, Aleksandar and Schmidt, Ludwig and Tsipras, Dimitris and Vladu, Adrian},
  journal={arXiv preprint arXiv:1706.06083},
  year={2017}
}

@inproceedings{wu20244d,
  title={4d gaussian splatting for real-time dynamic scene rendering},
  author={Wu, Guanjun and Yi, Taoran and Fang, Jiemin and Xie, Lingxi and Zhang, Xiaopeng and Wei, Wei and Liu, Wenyu and Tian, Qi and Wang, Xinggang},
  booktitle={Proceedings of the IEEE/CVF conference on computer vision and pattern recognition},
  pages={20310--20320},
  year={2024}
}

@inproceedings{yang2024deformable,
  title={Deformable 3d gaussians for high-fidelity monocular dynamic scene reconstruction},
  author={Yang, Ziyi and Gao, Xinyu and Zhou, Wen and Jiao, Shaohui and Zhang, Yuqing and Jin, Xiaogang},
  booktitle={Proceedings of the IEEE/CVF conference on computer vision and pattern recognition},
  pages={20331--20341},
  year={2024}
}

@article{wu2024opengaussian,
  title={Opengaussian: Towards point-level 3d gaussian-based open vocabulary understanding},
  author={Wu, Yanmin and Meng, Jiarui and Li, Haijie and Wu, Chenming and Shi, Yahao and Cheng, Xinhua and Zhao, Chen and Feng, Haocheng and Ding, Errui and Wang, Jingdong and others},
  journal={Advances in Neural Information Processing Systems},
  volume={37},
  pages={19114--19138},
  year={2024}
}

@inproceedings{ye2024gaussian,
  title={Gaussian grouping: Segment and edit anything in 3d scenes},
  author={Ye, Mingqiao and Danelljan, Martin and Yu, Fisher and Ke, Lei},
  booktitle={European conference on computer vision},
  pages={162--179},
  year={2024},
  organization={Springer}
}

@inproceedings{shi2024language,
  title={Language embedded 3d gaussians for open-vocabulary scene understanding},
  author={Shi, Jin-Chuan and Wang, Miao and Duan, Hao-Bin and Guan, Shao-Hua},
  booktitle={Proceedings of the IEEE/CVF Conference on Computer Vision and Pattern Recognition},
  pages={5333--5343},
  year={2024}
}

@article{lu2025segment,
  title={Segment then splat: A unified approach for 3d open-vocabulary segmentation based on gaussian splatting},
  author={Lu, Yiren and Zhou, Yunlai and Qiao, Yiran and Song, Chaoda and Liang, Tuo and Ma, Jing and Yin, Yu},
  journal={arXiv preprint arXiv:2503.22204},
  year={2025}
}

@article{goodfellow2014explaining,
  title={Explaining and harnessing adversarial examples},
  author={Goodfellow, Ian J and Shlens, Jonathon and Szegedy, Christian},
  journal={arXiv preprint arXiv:1412.6572},
  year={2014}
}

@inproceedings{moosavi2016deepfool,
  title={Deepfool: a simple and accurate method to fool deep neural networks},
  author={Moosavi-Dezfooli, Seyed-Mohsen and Fawzi, Alhussein and Frossard, Pascal},
  booktitle={Proceedings of the IEEE conference on computer vision and pattern recognition},
  pages={2574--2582},
  year={2016}
}

@inproceedings{andriushchenko2020square,
  title={Square attack: a query-efficient black-box adversarial attack via random search},
  author={Andriushchenko, Maksym and Croce, Francesco and Flammarion, Nicolas and Hein, Matthias},
  booktitle={European conference on computer vision},
  pages={484--501},
  year={2020},
  organization={Springer}
}

@article{liu2016delving,
  title={Delving into transferable adversarial examples and black-box attacks},
  author={Liu, Yanpei and Chen, Xinyun and Liu, Chang and Song, Dawn},
  journal={arXiv preprint arXiv:1611.02770},
  year={2016}
}

@article{qiao2025counterfactual,
  title={Counterfactual Visual Explanation via Causally-Guided Adversarial Steering},
  author={Qiao, Yiran and Liu, Disheng and Lu, Yiren and Yin, Yu and Du, Mengnan and Ma, Jing},
  journal={arXiv preprint arXiv:2507.09881},
  year={2025}
}

@article{zhang2021causaladv,
  title={Causaladv: Adversarial robustness through the lens of causality},
  author={Zhang, Yonggang and Gong, Mingming and Liu, Tongliang and Niu, Gang and Tian, Xinmei and Han, Bo and Sch{\"o}lkopf, Bernhard and Zhang, Kun},
  journal={arXiv preprint arXiv:2106.06196},
  year={2021}
}

@inproceedings{wu2025difix3d+,
  title={Difix3d+: Improving 3d reconstructions with single-step diffusion models},
  author={Wu, Jay Zhangjie and Zhang, Yuxuan and Turki, Haithem and Ren, Xuanchi and Gao, Jun and Shou, Mike Zheng and Fidler, Sanja and Gojcic, Zan and Ling, Huan},
  booktitle={Proceedings of the Computer Vision and Pattern Recognition Conference},
  pages={26024--26035},
  year={2025}
}

@inproceedings{jiang2021focal,
  title={Focal frequency loss for image reconstruction and synthesis},
  author={Jiang, Liming and Dai, Bo and Wu, Wayne and Loy, Chen Change},
  booktitle={Proceedings of the IEEE/CVF international conference on computer vision},
  pages={13919--13929},
  year={2021}
}

@article{rao2021global,
  title={Global filter networks for image classification},
  author={Rao, Yongming and Zhao, Wenliang and Zhu, Zheng and Lu, Jiwen and Zhou, Jie},
  journal={Advances in neural information processing systems},
  volume={34},
  pages={980--993},
  year={2021}
}

@inproceedings{zhang2024fregs,
  title={Fregs: 3d gaussian splatting with progressive frequency regularization},
  author={Zhang, Jiahui and Zhan, Fangneng and Xu, Muyu and Lu, Shijian and Xing, Eric},
  booktitle={Proceedings of the IEEE/CVF Conference on Computer Vision and Pattern Recognition},
  pages={21424--21433},
  year={2024}
}

@inproceedings{jang20253d,
  title={3d-gsw: 3d gaussian splatting for robust watermarking},
  author={Jang, Youngdong and Park, Hyunje and Yang, Feng and Ko, Heeju and Choo, Euijin and Kim, Sangpil},
  booktitle={Proceedings of the Computer Vision and Pattern Recognition Conference},
  pages={5938--5948},
  year={2025}
}

@inproceedings{xu2023wavenerf,
  title={Wavenerf: Wavelet-based generalizable neural radiance fields},
  author={Xu, Muyu and Zhan, Fangneng and Zhang, Jiahui and Yu, Yingchen and Zhang, Xiaoqin and Theobalt, Christian and Shao, Ling and Lu, Shijian},
  booktitle={Proceedings of the IEEE/CVF International Conference on Computer Vision},
  pages={18195--18204},
  year={2023}
}

@inproceedings{lou2024darenerf,
  title={Darenerf: Direction-aware representation for dynamic scenes},
  author={Lou, Ange and Planche, Benjamin and Gao, Zhongpai and Li, Yamin and Luan, Tianyu and Ding, Hao and Chen, Terrence and Noble, Jack and Wu, Ziyan},
  booktitle={Proceedings of the IEEE/CVF Conference on Computer Vision and Pattern Recognition},
  pages={5031--5042},
  year={2024}
}

@inproceedings{jang2024waterf,
  title={Waterf: Robust watermarks in radiance fields for protection of copyrights},
  author={Jang, Youngdong and Lee, Dong In and Jang, MinHyuk and Kim, Jong Wook and Yang, Feng and Kim, Sangpil},
  booktitle={Proceedings of the IEEE/CVF Conference on Computer Vision and Pattern Recognition},
  pages={12087--12097},
  year={2024}
}

@book{gonzalez2009digital,
  title={Digital image processing},
  author={Gonzalez, Rafael C},
  year={2009},
  publisher={Pearson education india}
}

@article{morelli2024_deep_image_matching,
  AUTHOR = {Morelli, L. and Ioli, F. and Maiwald, F. and Mazzacca, G. and Menna, F. and Remondino, F.},
  TITLE = {DEEP-IMAGE-MATCHING: A TOOLBOX FOR MULTIVIEW IMAGE MATCHING OF COMPLEX SCENARIOS},
  JOURNAL = {The International Archives of the Photogrammetry, Remote Sensing and Spatial Information Sciences},
  VOLUME = {XLVIII-2/W4-2024},
  YEAR = {2024},
  PAGES = {309--316},
  DOI = {10.5194/isprs-archives-XLVIII-2-W4-2024-309-2024}
}

@article{morelli2022photogrammetry,
  title={PHOTOGRAMMETRY NOW AND THEN--FROM HAND-CRAFTED TO DEEP-LEARNING TIE POINTS--},
  author={Morelli, Luca and Bellavia, Fabio and Menna, Fabio and Remondino, Fabio},
  journal={The International Archives of the Photogrammetry, Remote Sensing and Spatial Information Sciences},
  volume={48},
  pages={163--170},
  year={2022},
  publisher={Copernicus GmbH}
}

@article{ioli2024,
  title={Deep Learning Low-cost Photogrammetry for 4D Short-term Glacier
Dynamics Monitoring},
  author={Ioli, Francesco and Dematteis, Nicolò and Giordan, Daniele and Nex, Francesco and Pinto Livio},
  journal={PFG – Journal of Photogrammetry, Remote Sensing and Geoinformation Science},
  year={2024},
  DOI = {10.1007/s41064-023-00272-w}
}

@article{lin1973effective,
  title={An effective heuristic algorithm for the traveling-salesman problem},
  author={Lin, Shen and Kernighan, Brian W},
  journal={Operations research},
  volume={21},
  number={2},
  pages={498--516},
  year={1973},
  publisher={Informs}
}

@inproceedings{detone2018superpoint,
  title={Superpoint: Self-supervised interest point detection and description},
  author={DeTone, Daniel and Malisiewicz, Tomasz and Rabinovich, Andrew},
  booktitle={Proceedings of the IEEE conference on computer vision and pattern recognition workshops},
  pages={224--236},
  year={2018}
}

@inproceedings{lindenberger2023lightglue,
  title={Lightglue: Local feature matching at light speed},
  author={Lindenberger, Philipp and Sarlin, Paul-Edouard and Pollefeys, Marc},
  booktitle={Proceedings of the IEEE/CVF international conference on computer vision},
  pages={17627--17638},
  year={2023}
}

@inproceedings{barron2022mip,
  title={Mip-nerf 360: Unbounded anti-aliased neural radiance fields},
  author={Barron, Jonathan T and Mildenhall, Ben and Verbin, Dor and Srinivasan, Pratul P and Hedman, Peter},
  booktitle={Proceedings of the IEEE/CVF conference on computer vision and pattern recognition},
  pages={5470--5479},
  year={2022}
}

@article{Knapitsch2017,
    author    = {Arno Knapitsch and Jaesik Park and Qian-Yi Zhou and Vladlen Koltun},
    title     = {Tanks and Temples: Benchmarking Large-Scale Scene Reconstruction},
    journal   = {ACM Transactions on Graphics},
    volume    = {36},
    number    = {4},
    year      = {2017},
}

@article{mildenhall2019local,
  title={Local light field fusion: Practical view synthesis with prescriptive sampling guidelines},
  author={Mildenhall, Ben and Srinivasan, Pratul P and Ortiz-Cayon, Rodrigo and Kalantari, Nima Khademi and Ramamoorthi, Ravi and Ng, Ren and Kar, Abhishek},
  journal={ACM Transactions on Graphics (ToG)},
  volume={38},
  number={4},
  pages={1--14},
  year={2019},
  publisher={ACM New York, NY, USA}
}

@inproceedings{lee2024compact,
  title={Compact 3d gaussian representation for radiance field},
  author={Lee, Joo Chan and Rho, Daniel and Sun, Xiangyu and Ko, Jong Hwan and Park, Eunbyung},
  booktitle={Proceedings of the IEEE/CVF Conference on Computer Vision and Pattern Recognition},
  pages={21719--21728},
  year={2024}
}

@article{wang2004image,
  title={Image quality assessment: from error visibility to structural similarity},
  author={Wang, Zhou and Bovik, Alan C and Sheikh, Hamid R and Simoncelli, Eero P},
  journal={IEEE transactions on image processing},
  volume={13},
  number={4},
  pages={600--612},
  year={2004},
  publisher={IEEE}
}

@article{li2025remedygs,
  title={RemedyGS: Defend 3D Gaussian Splatting against Computation Cost Attacks},
  author={Li, Yanping and Liu, Zhening and Li, Zijian and Lin, Zehong and Zhang, Jun},
  journal={arXiv preprint arXiv:2511.22147},
  year={2025}
}

\clearpage
\appendix
\begin{center}
{\LARGE\textbf{Appendix}}
\end{center}
\section{Additional Implementation Details}
\label{sec:AID}
\subsection{Scene Selection.} For the Mip-NeRF 360 dataset, we use all available scenes: \textit{bicycle}, \textit{bonsai}, \textit{counter}, \textit{flowers}, \textit{garden}, \textit{kitchen}, \textit{room}, \textit{stump}, and \textit{treehill}. Eight static scenes are chosen from the Tanks-and-Temples dataset for evaluation: \textit{Auditorium}, \textit{Ballroom}, \textit{Caterpillar}, \textit{Courtroom}, \textit{M60}, \textit{Meetingroom}, \textit{Train}, and \textit{Truck}. We select the following scenes from the LLFF dataset: \textit{fern}, \textit{flower}, \textit{fortress}, \textit{horns}, \textit{orchids}, \textit{room}, and \textit{trex}.

\subsection{Poisoned Data Preparation.} The perturbed version of the Mip-NeRF 360 dataset is provided in Poison-Splat \cite{lu2024poison}, and thus we directly adopt it in our experiments. For the Tanks-and-Temples dataset, we generate the corresponding perturbed version using the script provided by Poison-Splat. The attack strength is set to $\epsilon = 16/255$, with 36,000 attack iterations. To evaluate the effectiveness of DefenseSplat under varying attack intensities, we additionally consider attack strengths of $\epsilon=32/255$ and $\epsilon=64/255$. For the LLFF dataset, to reduce memory consumption, we construct the adversarial data using quarter-resolution images. The attack strength is still set to $\epsilon=16/255$, with 30,000 attack iterations. It is important to note that in the Poison-Splat setting, the attack is applied only to the multi-view images, while the COLMAP-generated sparse reconstruction (including camera poses and intrinsics) remains unchanged. Therefore, during training on the adversarial images, we still use the camera data obtained from the clean images. For the LLFF dataset, the provided camera intrinsics are associated with a \textit{simple radial} camera model, which is incompatible with 3DGS reconstruction. Therefore, before generating the adversarial images, we manually re-run COLMAP to obtain camera intrinsics and poses under the pinhole model, along with the corresponding undistorted images.

\begin{table}
\centering
\caption{Quantitative results of deep image matching on different frequency components. (CR represents matching rate for clean input, PR means matching rate for poisoned input.)}
\begin{tabular}{ccccc}
\toprule
\multirow{2}{*}{Scene} & \multicolumn{2}{c}{CR(\%)} & \multicolumn{2}{c}{PR(\%)} \\ \cline{2-5} 
                       & LL          & LH\&HL       & LL          & LH\&HL       \\ \hline
\multicolumn{5}{c}{\cellcolor{gray!20}\textit{Mip-NeRF 360}}                                                 \\ \hline
bicycle                & 59.21       & 54.02        & 56.18       & 42.96        \\
bonsai                 & 34.76       & 26.99        & 33.17       & 21.94        \\
counter                & 60.63       & 55.33        & 58.26       & 46.35        \\
flowers                & 35.44       & 24.97        & 35.34       & 20.87        \\
garden                 & 52.61       & 32.46        & 51.32       & 24.74        \\
kitchen                & 54.78       & 48.47        & 54.29       & 38.52        \\
room                   & 53.36       & 50.02        & 50.1        & 40.11        \\
stump                  & 33.00       & 27.80        & 25.04       & 20.83        \\
treehill               & 30.62       & 20.50        & 28.92       & 17.69        \\
\rowcolor{cyan!15}
average                & 46.05       & 37.84        & 43.62       & 30.44        \\ \hline
\multicolumn{5}{c}{\cellcolor{gray!20}\textit{Tanks-and-Temples}}                                            \\ \hline
Auditorium             & 49.69       & 48.38        & 45.66       & 35.48        \\
Ballroom               & 59.37       & 51.10        & 57.75       & 43.68        \\
Caterpillar            & 59.75       & 49.45        & 58.45       & 46.81        \\
Courtroom              & 51.30       & 44.31        & 48.50       & 37.27        \\
M60                    & 57.67       & 52.83        & 56.41       & 45.15        \\
Meetingroom            & 54.43       & 51.68        & 52.88       & 42.66        \\
Train                  & 55.15       & 48.26        & 53.81       & 41.90        \\
Truck                  & 53.72       & 46.46        & 52.49       & 39.71        \\
\rowcolor{cyan!15}
average                & 55.14       & 49.06        & 53.24       & 41.58        \\ \hline
\multicolumn{5}{c}{\cellcolor{gray!20}\textit{LLFF}}                                                         \\ \hline
fern                   & 47.95       & 44.50        & 46.78       & 40.43        \\
flower                 & 69.39       & 63.70        & 68.19       & 55.66        \\
fortress               & 62.51       & 48.82        & 57.88       & 42.59        \\
horns                  & 58.58       & 51.09        & 57.94       & 47.73        \\
orchids                & 45.74       & 37.90        & 45.60       & 36.87        \\
room                   & 61.68       & 57.28        & 56.37       & 46.54        \\
trex                   & 59.90       & 52.03        & 58.61       & 47.12        \\
\rowcolor{cyan!15}
average                & 57.96       & 50.76        & 55.91       & 45.27        \\ \bottomrule
\end{tabular}
\label{tab4}
\end{table}

\section{Additional Experimental Results}
\subsection{Frequency-Aware Analysis of Adversarial Perturbations}
Since the energy proportion of the $HH$ subband is negligible and diagonal textures are relatively rare, we omit analysis of the $HH$ component. Instead, we average the results of the $LH$ and $HL$ subbands to represent the high-frequency content. As shown in \cref{tab4}, all three datasets exhibit the same trend: the decline in image matching after attack is more pronounced in the high-frequency components than in the low-frequency components. Specifically, for Mip-NeRF 360, the matching rate in the low-frequency band drops by 5.28\%, while the high-frequency band drops by 19.56\%; for Tanks-and-Temples, the drops are 3.45\% and 15.25\%, respectively; and for LLFF, the corresponding values are 3.54\% and 10.82\%.

\begin{table}
\centering
\caption{Quantitative results of deep image matching on different frequency components over different attack strengths (Tanks-and-Temples dataset).}
\begin{tabular}{ccccc}
\toprule
\multirow{2}{*}{$\epsilon$} & \multicolumn{2}{c}{CR(\%)} & \multicolumn{2}{c}{PR(\%)} \\ \cline{2-5} 
                   & LL          & LH\&HL       & LL          & LH\&HL       \\ \hline
16/255             & 54.44       & 47.36        & 53.15       & 40.81        \\
32/255             & 54.44       & 47.36        & 51.23       & 37.35        \\
64/255             & 54.44       & 47.36        & 43.93       & 35.23        \\ \bottomrule
\end{tabular}
\label{tab5}
\end{table}

\begin{table*}[th]
\centering
\caption{Quantitative comparison on {LLFF}. Colors indicate the \shadeText{red}{best} and \shadeText{orange}{second best}
results. \#G/M indicates the number of Gaussians (Million).}
{\fontsize{8pt}{9pt}\selectfont
\begin{tabular}{l|ccc|cccc}
\toprule
Method & Training Time $\downarrow$  & \#G/M $\downarrow$ & GPU Mem (MB) $\downarrow$ & PSNR $\uparrow$ & SSIM $\uparrow$ & LPIPS $\downarrow$ & FPS $\uparrow$ \\
\midrule
3DGS         & 0:24:47 & 2.03 & 5407 & 25.04 & 0.5856 & 0.4344 & 196 \\
Compact GS   & 0:46:01 & 1.75 & 8231 & 24.07 & 0.5540 & 0.4453 & 232 \\
Difix        & 0:16:11 & 0.99 & 3360 & 25.09 & 0.7968 & 0.2654 & 437 \\
\midrule
Ours         & \cellcolor[HTML]{F9CB9C}0:12:01 & \cellcolor[HTML]{F9CB9C}0.54 & \cellcolor[HTML]{F9CB9C}2919 & \cellcolor[HTML]{EA9999}27.67 & \cellcolor[HTML]{F9CB9C}0.8620 & \cellcolor[HTML]{F9CB9C}0.2480 & \cellcolor[HTML]{F9CB9C}749 \\
Ours+ReLU    & \cellcolor[HTML]{EA9999}0:11:47 & \cellcolor[HTML]{EA9999}0.53 & \cellcolor[HTML]{EA9999}2144 & \cellcolor[HTML]{F9CB9C}27.63 & \cellcolor[HTML]{EA9999}0.8635 & \cellcolor[HTML]{EA9999}0.2461 & \cellcolor[HTML]{EA9999}813 \\
\bottomrule
\end{tabular}}
\label{tab6}
\end{table*}


\begin{table}[t]
\centering
\caption{Quantitative results of our method on clean inputs (LLFF).}
\resizebox{\linewidth}{!}{
\begin{tabular}{l|ccc|cccc}
\toprule
Method & Training Time $\downarrow$ & \#G/M $\downarrow$ & GPU Mem (MB) $\downarrow$ & PSNR $\uparrow$ & SSIM $\uparrow$ & LPIPS $\downarrow$ & FPS $\uparrow$ \\
\midrule
Clean               & 0:14:05 & 0.69 & 3162 & 31.29 & 0.9431 & 0.1224  & 650 \\
Clean + Ours        & 0:11:58 & 0.50 & 3130 & 28.39 & 0.8919 & 0.1992  & 841 \\
Clean + Ours + ReLU & 0:11:47 & 0.48 & 2161 & 28.34 & 0.8910 & 0.1979  & 886 \\
\bottomrule
\end{tabular}
}
\label{tab7}
\end{table}

We also analyze the changes in image matching rates under different attack intensities. Specifically, we use the Train and Truck scenes from the Tanks-and-Temples dataset, as these are among the most commonly used scenes in prior work. As shown in \cref{tab5}, when $\epsilon = 16/255$, the low-frequency matching rate drops by 2.37\%, while the high-frequency matching rate drops by 13.83\%. At $\epsilon = 32/255$, this degradation increases to 5.90\% and 21.14\%, respectively. When $\epsilon = 64/255$, the corresponding values are 19.31\% and 25.61\%. It can be observed that as the attack strength increases, the target of the attack gradually shifts from high-frequency to low-frequency components. It is reasonable to predict that as the attack strength further increases, the degradation in low-frequency matching may even exceed that of high-frequency. However, this observation does not undermine the effectiveness of our defense strategy. The first reason is that adversarial perturbations are typically expected to remain imperceptible to the human eye, rendering excessively strong attacks of little practical relevance in real-world scenarios. The second reason is that even under stronger attack intensities, high-frequency components are still targeted; therefore, filtering out high-frequency information can still effectively remove part of the perturbation. As a result, our defense strategy remains effective.

\subsection{Additional Quantitative Results}
We provide more quantitative results on the LLFF dataset in this section, as shown in \cref{tab6} and \cref{tab7}. Our method ranks among the top two across all evaluation metrics. Since the previous quantitative results only report the average performance on each dataset, we provide detailed per-scene results for each dataset from \cref{tab8} to \cref{tab10_1}. In certain scenes, our method’s LPIPS score is slightly inferior to that of Difix3D+. This may be attributed to the fact that Difix3D+ employs a diffusion model as its backbone. The denoising capability of the pre-trained diffusion model, along with the rich priors it encodes, may enable it to produce outputs that are visually closer to the clean images when the inputs are adversarially attacked. However, for 3D reconstruction tasks, pixel-level consistency between the generated and ground truth images is more critical. Our method consistently outperforms all baselines across all scenes in terms of PSNR and SSIM, demonstrating its superiority. It is important to note that the primary objective of our task is to enhance the robustness of 3DGS, specifically by reducing the rate of training failure on the server side. Because if excessive computational resources are consumed, training may be interrupted at any time or even fail to start, not to mention the rendering quality in the subsequent stages. As shown by the results, our method is capable of completing training for each scene within a shorter time duration, using a smaller number of Gaussians and minimal GPU memory. While achieving this objective, it is widely agreed that higher rendering quality is always preferable. Our method successfully meets this requirement by delivering the highest average rendering quality among all methods, while maintaining high efficiency.

\subsection{Additional Qualitative Results}
\begin{figure*}
    \centering
    \includegraphics[width=0.9\textwidth]{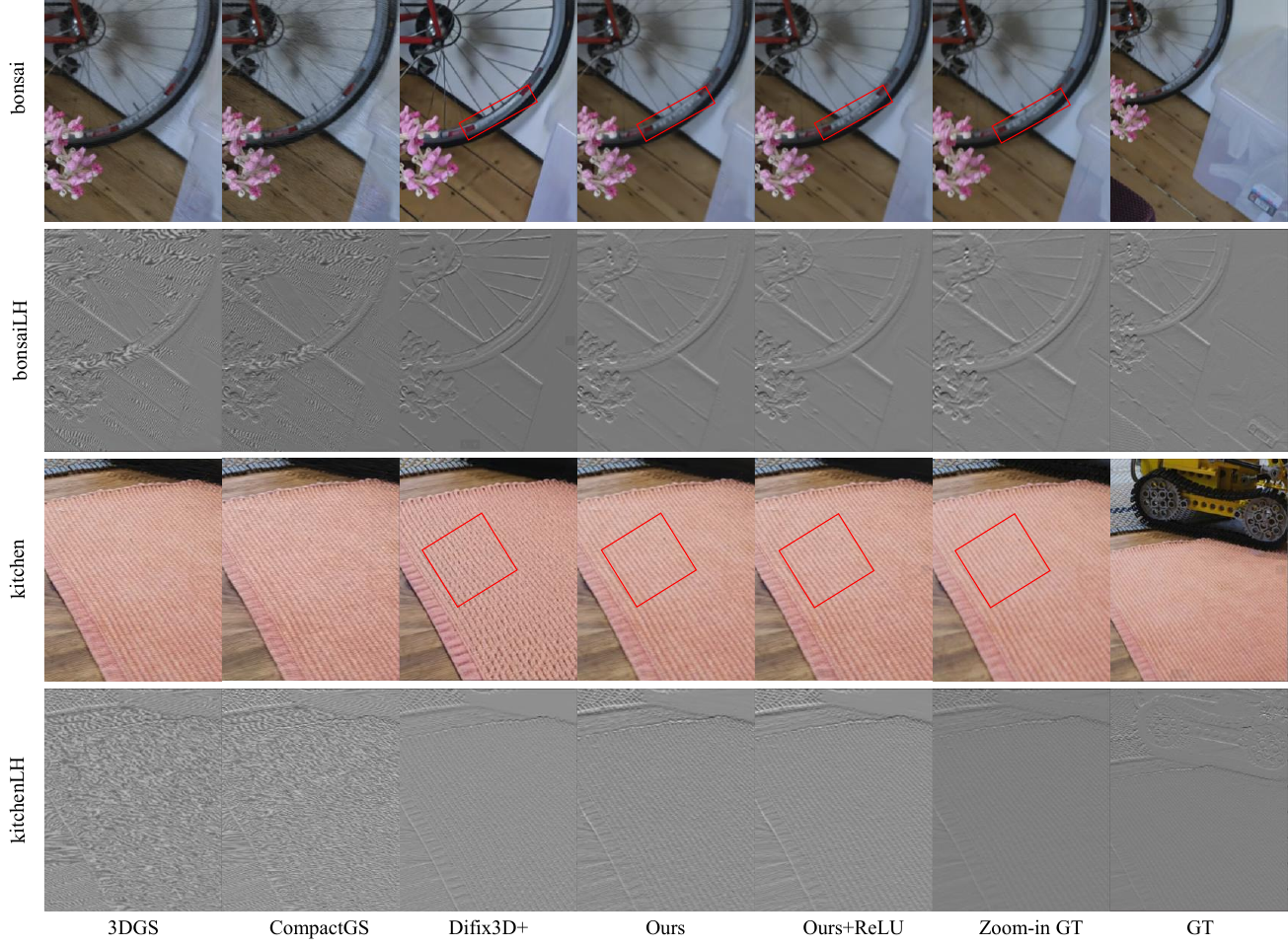} 
    \vspace{-0.5em}
    \caption{Comparison of reconstruction quality of all methods and ground truth (GT) on Mip-NeRF 360 datasets (bonsai \& kitchen).}
    \label{fig5}
    \vspace{-4mm}
\end{figure*}

\begin{figure*}
    \centering
    \includegraphics[width=0.9\textwidth]{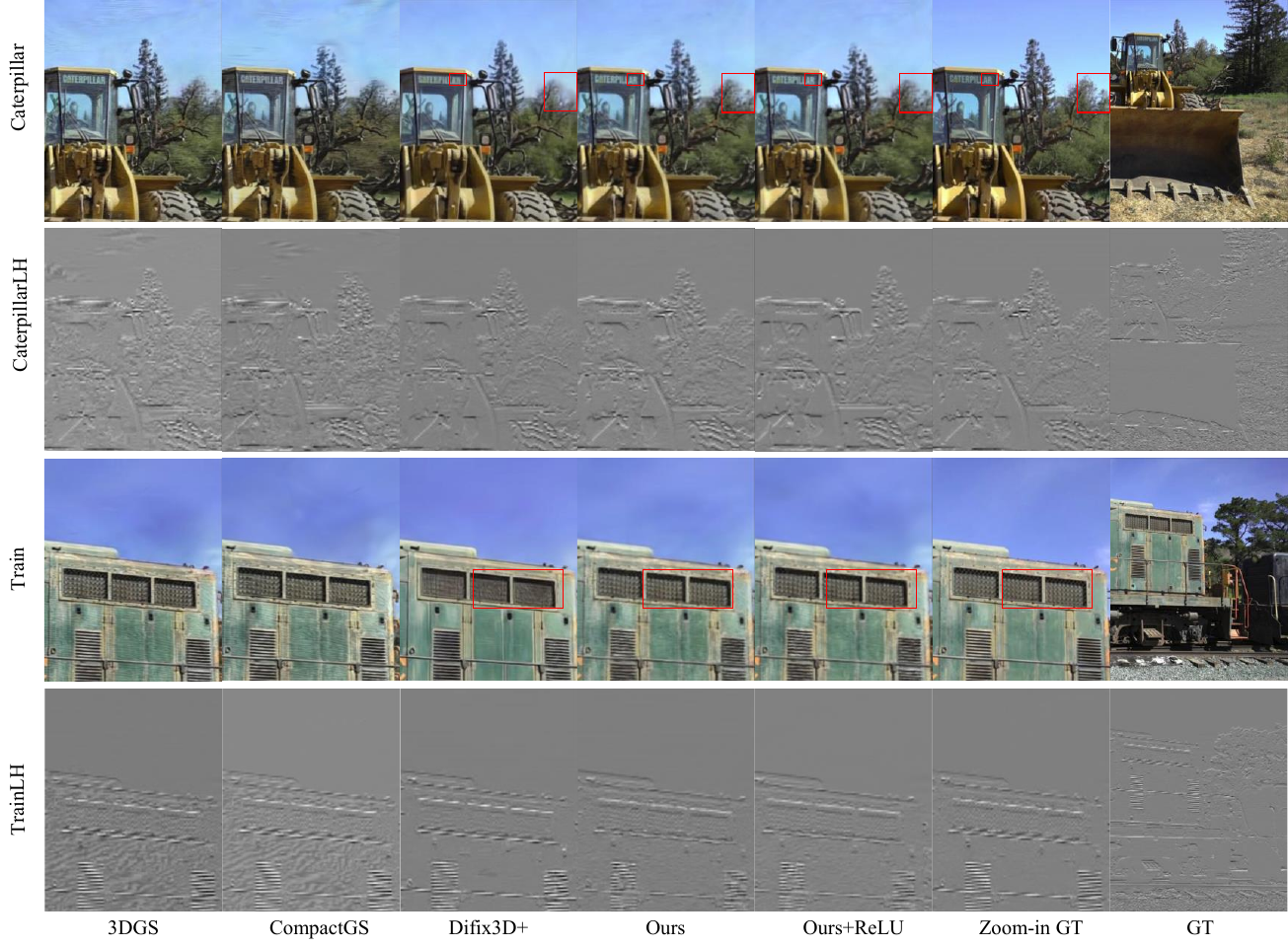} 
    \vspace{-0.5em}
    \caption{Comparison of reconstruction quality of all methods and ground truth (GT) on Tanks-and-Temples datasets (Caterpillar \& Train).}
    \label{fig6}
    \vspace{-4mm}
\end{figure*}

\begin{figure*}
    \centering
    \includegraphics[width=0.9\textwidth]{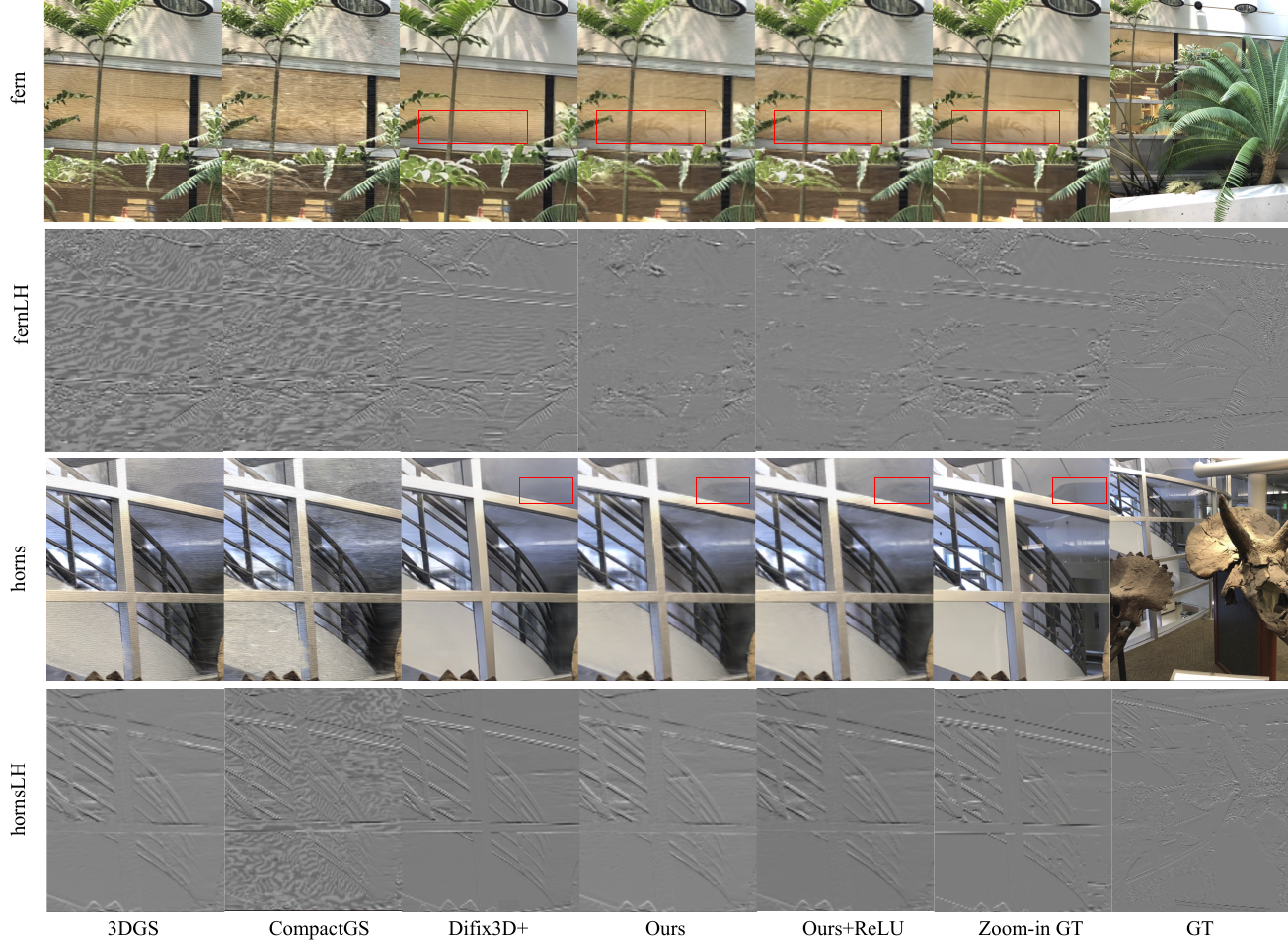} 
    \vspace{-0.5em}
    \caption{Comparison of reconstruction quality of all methods and ground truth (GT) on LLFF datasets (fern \& horns).}
    \label{fig7}
    \vspace{-4mm}
\end{figure*}

\looseness -1
\cref{fig5}, \cref{fig6}, and \cref{fig7} present additional rendering results. It is evident that the rendered images from the original 3DGS and CompactGS exhibit significant adversarial artifacts, indicating that these two methods are nearly ineffective at defending against 3D adversarial attacks. Difix3D+ improves the defense performance to some extent, but it also introduces several issues. In certain scenes, Difix3D+ excessively smooths the images to enhance quality, which results in the loss of fine details; in other cases, it sharpens the entire image or even slightly alters the color of some objects. Additionally, Difix3D+ sometimes introduces artificial textures that were not present in the original images. This result is expected, as Difix3D+ was originally trained on images with poor rendering quality due to sparse reconstruction. However, the distribution of these images differs from that of adversarially perturbed ones. Consequently, when applying single-step diffusion inference to perturbed images, the performance is naturally less satisfactory. The visualization of the $LH$ subband also reflects the effectiveness of our defense. Since $LH$ highlights the presence of perturbation textures more clearly, the reduction or even disappearance of such textures provides additional evidence for the success of our defense strategy.

\begin{table*}
\centering
\caption{Detailed quantitative comparison on {Mip-NeRF 360}. Colors indicate the \shadeText{red}{best} and \shadeText{orange}{second best}
results. \#G/M indicates the number of Gaussians (Million).}
{\fontsize{8pt}{9pt}\selectfont
\begin{tabular}{l|ccc|cccc}
\toprule
Method & Training Time $\downarrow$  & \#G/M $\downarrow$ & GPU Mem (MB) $\downarrow$ & PSNR $\uparrow$ & SSIM $\uparrow$ & LPIPS $\downarrow$ & FPS $\uparrow$ \\
\midrule
\multicolumn{8}{c}{\cellcolor{gray!20}\textit{bicycle}} \\
\midrule
3DGS         & 1:12:03 & 8.16 & 23704 & 23.88 & 0.6630 & 0.4038 & 48 \\
Compact GS   & 1:41:07 & 4.22 & 23190 & 23.12 & 0.6305 & 0.4327 & 77 \\
Difix        & 0:57:23 & 5.98 & 19984 & 22.91 & 0.6753 & 0.3586 & 67 \\
\midrule
Ours         & \cellcolor[HTML]{F9CB9C}0:43:18 & \cellcolor[HTML]{F9CB9C} 3.92 & \cellcolor[HTML]{F9CB9C}13902 & \cellcolor[HTML]{EA9999}24.77 & \cellcolor[HTML]{EA9999}0.7606 & \cellcolor[HTML]{EA9999}0.3400 & \cellcolor[HTML]{F9CB9C}104 \\
Ours+ReLU    & \cellcolor[HTML]{EA9999}0:42:39 & \cellcolor[HTML]{EA9999}3.71 & \cellcolor[HTML]{EA9999}13420 & \cellcolor[HTML]{F9CB9C}24.66 & \cellcolor[HTML]{F9CB9C}0.7579 & \cellcolor[HTML]{F9CB9C}0.3442 & \cellcolor[HTML]{EA9999}112 \\
\midrule
\multicolumn{8}{c}{\cellcolor{gray!20}\textit{bonsai}} \\
\midrule
3DGS         & 0:47:37 & 4.88 & 18496 & 27.38 & 0.6435 & 0.5054 & 82 \\
Compact GS   & 1:09:08 & 2.20 & 18318 & 27.30 & 0.6587 & 0.5041 & 168 \\
Difix        & 0:31:58 & 1.89 & 12394  & 25.71 & 0.8281 & \cellcolor[HTML]{F9CB9C}0.3440 & 221 \\
\midrule
Ours         & \cellcolor[HTML]{EA9999}0:24:19 & \cellcolor[HTML]{F9CB9C}0.89 & \cellcolor[HTML]{F9CB9C}11494  & \cellcolor[HTML]{F9CB9C}30.78 & \cellcolor[HTML]{F9CB9C}0.8875 & 0.3384 & \cellcolor[HTML]{F9CB9C}488 \\
Ours+ReLU    & \cellcolor[HTML]{F9CB9C}0:24:31 & \cellcolor[HTML]{EA9999}0.86 & \cellcolor[HTML]{EA9999}10498  & \cellcolor[HTML]{EA9999}30.79 & \cellcolor[HTML]{EA9999}0.8910 & \cellcolor[HTML]{EA9999}0.3313 & \cellcolor[HTML]{EA9999}506 \\
\midrule
\multicolumn{8}{c}{\cellcolor{gray!20}\textit{counter}} \\
\midrule
3DGS         & 0:57:48 & 4.31 & 23320 & 26.23 & 0.6224 & 0.4821 & 89 \\
Compact GS   & 1:08:36 & 1.79 & 21608 & 26.30 & 0.6523 & 0.4819 & 203 \\
Difix        & 0:40:57 & 1.86 & 14720  & 26.39 & 0.8264 & \cellcolor[HTML]{EA9999}0.3413 & 216 \\
\midrule
Ours         & \cellcolor[HTML]{F9CB9C}0:31:12 & \cellcolor[HTML]{F9CB9C}1.05 & \cellcolor[HTML]{F9CB9C}11512  & \cellcolor[HTML]{F9CB9C}28.67 & \cellcolor[HTML]{F9CB9C}0.8470 & 0.3491 & \cellcolor[HTML]{F9CB9C}395 \\
Ours+ReLU    & \cellcolor[HTML]{EA9999}0:30:17 & \cellcolor[HTML]{EA9999}0.96 & \cellcolor[HTML]{EA9999}10600  & \cellcolor[HTML]{EA9999}28.89 & \cellcolor[HTML]{EA9999}0.8585 & \cellcolor[HTML]{F9CB9C}0.3470 & \cellcolor[HTML]{EA9999}437 \\
\midrule
\multicolumn{8}{c}{\cellcolor{gray!20}\textit{flowers}} \\
\midrule
3DGS         & 0:41:57 & 4.20 & 13872 & 22.53 & 0.6351 & 0.4284 & 96 \\
Compact GS   & 1:04:31 & 2.40 & 14412 & 21.81 & 0.5982 & 0.4557 & 154 \\
Difix        & 0:35:59 & 3.10 & 11572  & 21.74 & 0.6406 & 0.3936 & 132 \\
\midrule
Ours         & \cellcolor[HTML]{EA9999}0:29:37 & \cellcolor[HTML]{F9CB9C}2.17 & \cellcolor[HTML]{EA9999}9500  & \cellcolor[HTML]{F9CB9C}22.85 & \cellcolor[HTML]{F9CB9C}0.6857 & \cellcolor[HTML]{F9CB9C} 0.3915 & \cellcolor[HTML]{F9CB9C}193 \\
Ours+ReLU    & \cellcolor[HTML]{F9CB9C}0:29:41 & \cellcolor[HTML]{EA9999}2.08 & \cellcolor[HTML]{F9CB9C}9734  & \cellcolor[HTML]{EA9999}22.92 & \cellcolor[HTML]{EA9999}0.6886 & \cellcolor[HTML]{EA9999}0.3822 & \cellcolor[HTML]{EA9999}202 \\
\bottomrule
\end{tabular}}
\label{tab8}
\end{table*}

\begin{table*}
\centering
\caption{(Continued Table) Detailed quantitative comparison on {Mip-NeRF 360}. Colors indicate the \shadeText{red}{best} and \shadeText{orange}{second best}
results. \#G/M indicates the number of Gaussians (Million).}
{\fontsize{8pt}{9pt}\selectfont
\begin{tabular}{l|ccc|cccc}
\toprule
Method & Training Time $\downarrow$  & \#G/M $\downarrow$ & GPU Mem (MB) $\downarrow$ & PSNR $\uparrow$ & SSIM $\uparrow$ & LPIPS $\downarrow$ & FPS $\uparrow$ \\
\midrule
\multicolumn{8}{c}{\cellcolor{gray!20}\textit{garden}} \\
\midrule
3DGS         & 1:05:19 & 6.20 & 19246 & 25.00 & 0.6900 & 0.3731 & 62 \\
Compact GS   & 1:40:22 & 3.66 & 20806 & 24.93 & 0.6935 & 0.3858 & 97 \\
Difix        & 0:50:30 & 4.26 & 15226  & 24.88 & 0.736 & 0.3267 & 93 \\
\midrule
Ours         & \cellcolor[HTML]{F9CB9C}0:37:50 & \cellcolor[HTML]{F9CB9C}2.85 & \cellcolor[HTML]{F9CB9C}11804  & \cellcolor[HTML]{F9CB9C}26.88 & \cellcolor[HTML]{F9CB9C}0.8084 & \cellcolor[HTML]{F9CB9C} 0.2958 & \cellcolor[HTML]{F9CB9C}143 \\
Ours+ReLU    & \cellcolor[HTML]{EA9999}0:37:03 & \cellcolor[HTML]{EA9999}2.79 & \cellcolor[HTML]{EA9999}11450  & \cellcolor[HTML]{EA9999}27.26 & \cellcolor[HTML]{EA9999}0.8239 & \cellcolor[HTML]{EA9999}0.2714 & \cellcolor[HTML]{EA9999}148 \\
\midrule
\multicolumn{8}{c}{\cellcolor{gray!20}\textit{kitchen}} \\
\midrule
3DGS         & 1:13:39 & 5.89 & 23010 & 26.36 & 0.6620 & 0.3869 & 64 \\
Compact GS   & 1:33:36 & 3.37 & 24434 & 26.52 & 0.6854 & 0.3774 & 108 \\
Difix        & 0:56:06 & 3.12 & 19402  & 25.74 & 0.7998 & 0.2744 & 123 \\
\midrule
Ours         & \cellcolor[HTML]{F9CB9C}0:39:45 & \cellcolor[HTML]{F9CB9C}1.58 & \cellcolor[HTML]{F9CB9C}14056  & \cellcolor[HTML]{F9CB9C}29.80 & \cellcolor[HTML]{F9CB9C}0.8771 & \cellcolor[HTML]{F9CB9C} 0.2500 & \cellcolor[HTML]{F9CB9C}253 \\
Ours+ReLU    & \cellcolor[HTML]{EA9999}0:38:10 & \cellcolor[HTML]{EA9999}1.54 & \cellcolor[HTML]{EA9999}12912  & \cellcolor[HTML]{EA9999}30.27 & \cellcolor[HTML]{EA9999}0.8843 & \cellcolor[HTML]{EA9999}0.2388 & \cellcolor[HTML]{EA9999}262 \\
\midrule
\multicolumn{8}{c}{\cellcolor{gray!20}\textit{room}} \\
\midrule
3DGS         & 1:20:18 & 6.73 & 31902 & 26.94 & 0.5432 & 0.5143 & 58 \\
Compact GS   & 1:31:39 & 2.96 & 32784 & 27.30 & 0.5818 & 0.5100 & 122 \\
Difix        & 0:43:10 & 2.01 & 18130  & 27.56 & 0.8264 & \cellcolor[HTML]{EA9999}0.3517 & 208 \\
\midrule
Ours         & \cellcolor[HTML]{F9CB9C}0:32:42 & \cellcolor[HTML]{F9CB9C}1.23 & \cellcolor[HTML]{F9CB9C}18649  & \cellcolor[HTML]{F9CB9C}31.15 & \cellcolor[HTML]{F9CB9C}0.8329 & \cellcolor[HTML]{F9CB9C} 0.3598 & \cellcolor[HTML]{F9CB9C}348 \\
Ours+ReLU    & \cellcolor[HTML]{EA9999}0:30:53 & \cellcolor[HTML]{EA9999}1.15 & \cellcolor[HTML]{EA9999}12428  & \cellcolor[HTML]{EA9999}31.37 & \cellcolor[HTML]{EA9999}0.8432 & 0.3621 & \cellcolor[HTML]{EA9999}379 \\
\midrule
\multicolumn{8}{c}{\cellcolor{gray!20}\textit{stump}} \\
\midrule
3DGS         & 1:00:09 & 7.97 & 20388 & 25.72 & 0.6535 & 0.4681 & 51 \\
Compact GS   & 1:19:27 & 3.27 & 17658 & 25.16 & 0.6420 & 0.4863 & 112 \\
Difix        & 0:55:03 & 7.15 & 19428  & 23.84 & 0.6439 & 0.3975 & 57 \\
\midrule
Ours         & \cellcolor[HTML]{EA9999}0:37:06 & \cellcolor[HTML]{F9CB9C}3.93 & \cellcolor[HTML]{F9CB9C}12142  & \cellcolor[HTML]{F9CB9C}27.66 & \cellcolor[HTML]{F9CB9C}0.7837 & \cellcolor[HTML]{F9CB9C} 0.3686 & \cellcolor[HTML]{F9CB9C}107 \\
Ours+ReLU    & \cellcolor[HTML]{F9CB9C}0:37:32 & \cellcolor[HTML]{EA9999}3.80 & \cellcolor[HTML]{EA9999}11954  & \cellcolor[HTML]{EA9999}27.76 & \cellcolor[HTML]{EA9999}0.7853 & \cellcolor[HTML]{EA9999}0.3666 & \cellcolor[HTML]{EA9999}112 \\
\midrule
\multicolumn{8}{c}{\cellcolor{gray!20}\textit{treehill}} \\
\midrule
3DGS         & 0:50:22 & 4.87 & 14744 & 22.59 & 0.6079 & 0.4710 & 81 \\
Compact GS   & 1:13:24 & 3.17 & 16436 & 22.14 & 0.5941 & 0.4971 & 115 \\
Difix        & 0:40:26 & 3.51 & 12192  & 21.90 & 0.6170 & 0.4298 & 116 \\
\midrule
Ours         & \cellcolor[HTML]{F9CB9C}0:34:02 & \cellcolor[HTML]{F9CB9C}2.58 & \cellcolor[HTML]{F9CB9C}10042  & \cellcolor[HTML]{F9CB9C}23.28 & \cellcolor[HTML]{F9CB9C}0.6883 & \cellcolor[HTML]{EA9999} 0.4293 & \cellcolor[HTML]{F9CB9C}160 \\
Ours+ReLU    & \cellcolor[HTML]{EA9999}0:33:33 & \cellcolor[HTML]{EA9999}2.50 & \cellcolor[HTML]{EA9999}9608  & \cellcolor[HTML]{EA9999}23.28 & \cellcolor[HTML]{EA9999}0.6901 & \cellcolor[HTML]{F9CB9C}0.4319 & \cellcolor[HTML]{EA9999}168 \\
\bottomrule
\end{tabular}}
\label{tab8_1}
\end{table*}

\begin{table*}
\centering
\caption{Detailed quantitative comparison on {Tanks-and-Temples}. Colors indicate the \shadeText{red}{best} and \shadeText{orange}{second best}
results. \#G/M indicates the number of Gaussians (Million).}
{\fontsize{8pt}{9pt}\selectfont
\begin{tabular}{l|ccc|cccc}
\toprule
Method & Training Time $\downarrow$  & \#G/M $\downarrow$ & GPU Mem (MB) $\downarrow$ & PSNR $\uparrow$ & SSIM $\uparrow$ & LPIPS $\downarrow$ & FPS $\uparrow$ \\
\midrule
\multicolumn{8}{c}{\cellcolor{gray!20}\textit{Auditorium}} \\
\midrule
3DGS         & 0:29:45 & 2.42 & 11782 & 26.06 & 0.5697 & 0.5125 & 166 \\
Compact GS   & 0:43:13 & 1.79 & 11842 & 25.66 & 0.5924 & 0.5183 & 207 \\
Difix        & 0:17:57 & 0.76 & 5710 & 27.14 & 0.8713 & \cellcolor[HTML]{EA9999}0.3411 & 554 \\
\midrule
Ours         & \cellcolor[HTML]{EA9999}0:15:30 & \cellcolor[HTML]{F9CB9C} 0.50 & \cellcolor[HTML]{F9CB9C}5120 & \cellcolor[HTML]{F9CB9C}28.82 & \cellcolor[HTML]{F9CB9C}0.8682 &  \cellcolor[HTML]{F9CB9C}0.3761 & \cellcolor[HTML]{F9CB9C}877 \\
Ours+ReLU    & \cellcolor[HTML]{F9CB9C}0:16:00 & \cellcolor[HTML]{EA9999}0.46 & \cellcolor[HTML]{EA9999}4730 & \cellcolor[HTML]{EA9999}28.90 & \cellcolor[HTML]{EA9999}0.8779 &0.3790 & \cellcolor[HTML]{EA9999}951 \\
\midrule
\multicolumn{8}{c}{\cellcolor{gray!20}\textit{Ballroom}} \\
\midrule
3DGS         & 0:38:28 & 3.53 & 11362 & 24.68 & 0.6896 & 0.3679 & 112 \\
Compact GS   & 0:55:36 & 2.25 & 12426 & 24.26 & 0.6920 & 0.3746 & 164 \\
Difix        & 0:35:09 & 2.95 & 10598  & 24.12 & 0.7949 & \cellcolor[HTML]{EA9999}0.2642 & 135 \\
\midrule
Ours         & \cellcolor[HTML]{EA9999}0:25:13 & \cellcolor[HTML]{F9CB9C}1.79 & \cellcolor[HTML]{F9CB9C}7806  & \cellcolor[HTML]{F9CB9C}25.59 & \cellcolor[HTML]{F9CB9C}0.8250 & 0.2898 & \cellcolor[HTML]{F9CB9C}228 \\
Ours+ReLU    & \cellcolor[HTML]{F9CB9C}0:25:23 & \cellcolor[HTML]{EA9999}1.70 & \cellcolor[HTML]{EA9999}7090  & \cellcolor[HTML]{EA9999}25.77 & \cellcolor[HTML]{EA9999}0.8350 & \cellcolor[HTML]{F9CB9C}0.2747 & \cellcolor[HTML]{EA9999}244 \\
\midrule
\multicolumn{8}{c}{\cellcolor{gray!20}\textit{Caterpillar}} \\
\midrule
3DGS         & 0:22:45 & 1.76 & 7598 & 24.08 & 0.6793 & 0.4058 & 228 \\
Compact GS   & 0:30:03 & 1.13 & 7668 & 23.56 & 0.6870 & 0.4225 & 345 \\
Difix        & 0:18:52 & 1.18 & 6808  & 23.45 & 0.7575 & 0.3390 & 347 \\
\midrule
Ours         & \cellcolor[HTML]{EA9999}0:15:58 & \cellcolor[HTML]{F9CB9C}0.82 & \cellcolor[HTML]{F9CB9C}6000  & \cellcolor[HTML]{F9CB9C}24.97 & \cellcolor[HTML]{F9CB9C}0.7834 & \cellcolor[HTML]{F9CB9C}0.3341 & \cellcolor[HTML]{F9CB9C}513 \\
Ours+ReLU    & \cellcolor[HTML]{F9CB9C}0:16:22 & \cellcolor[HTML]{EA9999}0.79 & \cellcolor[HTML]{EA9999}5572  & \cellcolor[HTML]{EA9999}25.04 & \cellcolor[HTML]{EA9999}0.7928 & \cellcolor[HTML]{EA9999}0.3277 & \cellcolor[HTML]{EA9999}533 \\
\midrule
\multicolumn{8}{c}{\cellcolor{gray!20}\textit{Courtroom}} \\
\midrule
3DGS         & 0:41:12 & 4.64 & 13074 & 25.09 & 0.6266 & 0.4313 & 87 \\
Compact GS   & 1:00:36 & 2.52 & 14074 & 24.06 & 0.6315 & 0.4433 & 147 \\
Difix        & 0:30:24 & 2.77 & 9324  & 24.92 & 0.7968 & \cellcolor[HTML]{EA9999}0.3003 & 150 \\
\midrule
Ours         & \cellcolor[HTML]{EA9999}0:21:45 & \cellcolor[HTML]{F9CB9C}1.62 & \cellcolor[HTML]{F9CB9C}6854  & \cellcolor[HTML]{F9CB9C}26.11 & \cellcolor[HTML]{F9CB9C}0.8063 &  0.3448 & \cellcolor[HTML]{F9CB9C}265 \\
Ours+ReLU    & \cellcolor[HTML]{F9CB9C}0:22:20 & \cellcolor[HTML]{EA9999}1.55 & \cellcolor[HTML]{EA9999}6428  & \cellcolor[HTML]{EA9999}26.24 & \cellcolor[HTML]{EA9999}0.8152 & \cellcolor[HTML]{F9CB9C}0.3337 & \cellcolor[HTML]{EA9999}283 \\
\midrule
\multicolumn{8}{c}{\cellcolor{gray!20}\textit{M60}} \\
\midrule
3DGS         & 0:32:19 & 2.65 & 9978 & 25.62 & 0.6559 & 0.4030 & 149 \\
Compact GS   & 0:39:57 & 1.41 & 11830 & 25.50 & 0.6780 & 0.4010 & 272 \\
Difix        & 0:24:53 & 1.60 & 8000  & 25.27 & 0.8364 & 0.2724 & 258 \\
\midrule
Ours         & \cellcolor[HTML]{EA9999}0:20:12 & \cellcolor[HTML]{F9CB9C}0.95 & \cellcolor[HTML]{F9CB9C}6994  & \cellcolor[HTML]{F9CB9C}27.60 & \cellcolor[HTML]{F9CB9C}0.8635 & \cellcolor[HTML]{F9CB9C} 0.2739 & \cellcolor[HTML]{F9CB9C}447 \\
Ours+ReLU    & \cellcolor[HTML]{F9CB9C}0:20:20 & \cellcolor[HTML]{EA9999}0.91 & \cellcolor[HTML]{EA9999}6318  & \cellcolor[HTML]{EA9999}27.80 & \cellcolor[HTML]{EA9999}0.8717 & \cellcolor[HTML]{EA9999}0.2638 & \cellcolor[HTML]{EA9999}467 \\
\midrule
\multicolumn{8}{c}{\cellcolor{gray!20}\textit{Meetingroom}} \\
\midrule
3DGS         & 0:31:10 & 2.72 & 9798 & 25.47 & 0.5958 & 0.4822 & 147 \\
Compact GS   & 0:44:28 & 1.81 & 10670 & 24.78 & 0.6134 & 0.4915 & 206 \\
Difix        & 0:23:00 & 1.45 & 7072  & 25.90 & 0.8435 & \cellcolor[HTML]{EA9999}0.3117 & 289 \\
\midrule
Ours         & \cellcolor[HTML]{EA9999}0:17:25 & \cellcolor[HTML]{F9CB9C}0.83 & \cellcolor[HTML]{F9CB9C}6308  & \cellcolor[HTML]{F9CB9C}27.47 & \cellcolor[HTML]{F9CB9C}0.8474 &  0.3521 & \cellcolor[HTML]{F9CB9C}520 \\
Ours+ReLU    & \cellcolor[HTML]{F9CB9C}0:17:36 & \cellcolor[HTML]{EA9999}0.77 & \cellcolor[HTML]{EA9999}5712  & \cellcolor[HTML]{EA9999}27.58 & \cellcolor[HTML]{EA9999}0.8574 & \cellcolor[HTML]{F9CB9C}0.3464 & \cellcolor[HTML]{EA9999}564 \\
\bottomrule
\end{tabular}}
\label{tab9}
\end{table*}

\begin{table*}
\centering
\caption{(Continued Table) Detailed quantitative comparison on {Tanks-and-Temples}. Colors indicate the \shadeText{red}{best} and \shadeText{orange}{second best}
results. \#G/M indicates the number of Gaussians (Million).}
{\fontsize{8pt}{9pt}\selectfont
\begin{tabular}{l|ccc|cccc}
\toprule
Method & Training Time $\downarrow$  & \#G/M $\downarrow$ & GPU Mem (MB) $\downarrow$ & PSNR $\uparrow$ & SSIM $\uparrow$ & LPIPS $\downarrow$ & FPS $\uparrow$ \\
\midrule
\multicolumn{8}{c}{\cellcolor{gray!20}\textit{Train}} \\
\midrule
3DGS         & 0:21:40 & 1.30 & 7596 & 23.89 & 0.7231 & 0.3605 & 303 \\
Compact GS   & 0:26:19 & 0.90 & 7742 & 23.13 & 0.7237 & 0.3799 & 434 \\
Difix        & 0:20:38 & 1.09 & 7666 & 23.02 & 0.7702 & \cellcolor[HTML]{F9CB9C}0.3086 & 372 \\
\midrule
Ours         & \cellcolor[HTML]{EA9999}0:17:15 & \cellcolor[HTML]{F9CB9C} 0.73 & \cellcolor[HTML]{F9CB9C}5972 & \cellcolor[HTML]{F9CB9C}24.21 & \cellcolor[HTML]{F9CB9C}0.8011 &  0.3101 & \cellcolor[HTML]{F9CB9C}562 \\
Ours+ReLU    & \cellcolor[HTML]{F9CB9C}0:17:19 & \cellcolor[HTML]{EA9999}0.70 & \cellcolor[HTML]{EA9999}5588 & \cellcolor[HTML]{EA9999}24.32 & \cellcolor[HTML]{EA9999}0.8080 & \cellcolor[HTML]{EA9999}0.3056 & \cellcolor[HTML]{EA9999}594 \\
\midrule
\multicolumn{8}{c}{\cellcolor{gray!20}\textit{Truck}} \\
\midrule
3DGS         & 0:33:35 & 3.04 & 9520 & 24.17 & 0.6001 & 0.3884 & 128 \\
Compact GS   & 0:47:36 & 2.04 & 11190 & 24.04 & 0.6139 & 0.3910 & 179 \\
Difix        & 0:22:40 & 1.88 & 6810  & 24.16 & 0.8033 & \cellcolor[HTML]{F9CB9C}0.2874 & 219 \\
\midrule
Ours         & \cellcolor[HTML]{EA9999}0:18:13 & \cellcolor[HTML]{F9CB9C}1.26 & \cellcolor[HTML]{EA9999}5312  & \cellcolor[HTML]{F9CB9C}25.56 & \cellcolor[HTML]{F9CB9C}0.8296 & 0.2780 & \cellcolor[HTML]{F9CB9C}337 \\
Ours+ReLU    & \cellcolor[HTML]{F9CB9C}0:19:14 & \cellcolor[HTML]{EA9999}1.21 & \cellcolor[HTML]{F9CB9C}5422  & \cellcolor[HTML]{EA9999}25.74 & \cellcolor[HTML]{EA9999}0.8404 & \cellcolor[HTML]{EA9999}0.2753 & \cellcolor[HTML]{EA9999}353 \\

\bottomrule
\end{tabular}}
\label{tab9_1}
\end{table*}

\begin{table*}
\centering
\caption{Detailed quantitative comparison on {LLFF}. Colors indicate the \shadeText{red}{best} and \shadeText{orange}{second best}
results. \#G/M indicates the number of Gaussians (Million).}
{\fontsize{8pt}{9pt}\selectfont
\begin{tabular}{l|ccc|cccc}
\toprule
Method & Training Time $\downarrow$  & \#G/M $\downarrow$ & GPU Mem (MB) $\downarrow$ & PSNR $\uparrow$ & SSIM $\uparrow$ & LPIPS $\downarrow$ & FPS $\uparrow$ \\
\midrule
\multicolumn{8}{c}{\cellcolor{gray!20}\textit{fern}} \\
\midrule
3DGS         & 0:27:44 & 2.22 & 5588 & 24.95 & 0.5988 & 0.4141 & 166 \\
Compact GS   & 0:50:16 & 2.01 & 8228 & 23.01 & 0.5362 & 0.4382 & 189 \\
Difix        & 0:18:24 & 1.20 & 3202 & 23.76 & 0.7471 & 0.2996 & 314 \\
\midrule
Ours         & \cellcolor[HTML]{F9CB9C}0:13:23 & \cellcolor[HTML]{F9CB9C} 0.67 & \cellcolor[HTML]{F9CB9C}2806 & \cellcolor[HTML]{F9CB9C}25.80 & \cellcolor[HTML]{F9CB9C}0.8205 &  \cellcolor[HTML]{F9CB9C}0.2888 & \cellcolor[HTML]{F9CB9C}560 \\
Ours+ReLU    & \cellcolor[HTML]{EA9999}0:12:44 & \cellcolor[HTML]{EA9999}0.65 & \cellcolor[HTML]{EA9999}2064 & \cellcolor[HTML]{EA9999}25.92 & \cellcolor[HTML]{EA9999}0.8250 & \cellcolor[HTML]{EA9999}0.2872 & \cellcolor[HTML]{EA9999}596 \\
\midrule
\multicolumn{8}{c}{\cellcolor{gray!20}\textit{flower}} \\
\midrule
3DGS         & 0:18:46 & 1.41 & 3978 & 25.27 & 0.5702 & 0.4580 & 279 \\
Compact GS   & 0:34:54 & 1.10 & 5992 & 24.51 & 0.5424 & 0.4564 & 353 \\
Difix        & 0:14:17 & 0.99 & 2958  & 25.42 & 0.7766 & 0.2840 & 407 \\
\midrule
Ours         & \cellcolor[HTML]{EA9999}0:10:14 & \cellcolor[HTML]{F9CB9C}0.49 & \cellcolor[HTML]{EA9999}1834  & \cellcolor[HTML]{EA9999}29.15 & \cellcolor[HTML]{F9CB9C}0.8681 & \cellcolor[HTML]{F9CB9C}0.2179 & \cellcolor[HTML]{F9CB9C}838 \\
Ours+ReLU    & \cellcolor[HTML]{F9CB9C}0:10:29 & \cellcolor[HTML]{EA9999}0.49 & \cellcolor[HTML]{F9CB9C}1916  & \cellcolor[HTML]{F9CB9C}29.08 & \cellcolor[HTML]{EA9999}0.8695 & \cellcolor[HTML]{EA9999}0.2099 & \cellcolor[HTML]{EA9999}843 \\
\midrule
\multicolumn{8}{c}{\cellcolor{gray!20}\textit{fortress}} \\
\midrule
3DGS         & 0:26:39 & 2.22 & 5538 & 25.51 & 0.5405 & 0.4640 & 172 \\
Compact GS   & 1:01:14 & 2.32 & 10948 & 25.13 & 0.5225 & 0.4696 & 158 \\
Difix        & 0:18:25 & 1.09 & 3318  & 25.19 & 0.7264 & 0.2867 & 353 \\
\midrule
Ours         & \cellcolor[HTML]{EA9999}0:11:52 & \cellcolor[HTML]{F9CB9C}0.44 & \cellcolor[HTML]{F9CB9C}2614  & \cellcolor[HTML]{EA9999}30.69 & \cellcolor[HTML]{F9CB9C}0.8769 & \cellcolor[HTML]{EA9999}0.2408 & \cellcolor[HTML]{F9CB9C}711 \\
Ours+ReLU    & \cellcolor[HTML]{F9CB9C}0:12:10 & \cellcolor[HTML]{EA9999}0.44 & \cellcolor[HTML]{EA9999}1902  & \cellcolor[HTML]{F9CB9C}30.53 & \cellcolor[HTML]{EA9999}0.8771 & \cellcolor[HTML]{F9CB9C}0.2449 & \cellcolor[HTML]{EA9999}924 \\
\midrule
\multicolumn{8}{c}{\cellcolor{gray!20}\textit{horns}} \\
\midrule
3DGS         & 0:24:30 & 1.89 & 5392 & 24.90 & 0.6241 & 0.4049 & 204 \\
Compact GS   & 0:42:36 & 1.43 & 7358 & 24.23 & 0.5974 & 0.4171 & 261 \\
Difix        & 0:17:40 & 1.03 & \cellcolor[HTML]{F9CB9C}3408  & 25.24 & 0.7989 & 0.2759 & 387 \\
\midrule
Ours         & \cellcolor[HTML]{F9CB9C}0:13:28 & \cellcolor[HTML]{F9CB9C}0.58 & 3552  & \cellcolor[HTML]{EA9999}26.80 & \cellcolor[HTML]{F9CB9C}0.8464 & \cellcolor[HTML]{EA9999}0.2459 & \cellcolor[HTML]{F9CB9C}629 \\
Ours+ReLU    & \cellcolor[HTML]{EA9999}0:13:09 & \cellcolor[HTML]{EA9999}0.57 & \cellcolor[HTML]{EA9999}2684  & \cellcolor[HTML]{F9CB9C}26.79 & \cellcolor[HTML]{EA9999}0.8474 & \cellcolor[HTML]{F9CB9C}0.2465 & \cellcolor[HTML]{EA9999}709 \\

\bottomrule
\end{tabular}}
\label{tab10}
\end{table*}

\begin{table*}
\centering
\caption{(Continued Table) Detailed quantitative comparison on {LLFF}. Colors indicate the \shadeText{red}{best} and \shadeText{orange}{second best}
results. \#G/M indicates the number of Gaussians (Million).}
{\fontsize{8pt}{9pt}\selectfont
\begin{tabular}{l|ccc|cccc}
\toprule
Method & Training Time $\downarrow$  & \#G/M $\downarrow$ & GPU Mem (MB) $\downarrow$ & PSNR $\uparrow$ & SSIM $\uparrow$ & LPIPS $\downarrow$ & FPS $\uparrow$ \\
\midrule
\multicolumn{8}{c}{\cellcolor{gray!20}\textit{orchids}} \\
\midrule
3DGS         & 0:21:14 & 1.61 & 4142 & 23.77 & 0.7026 & 0.3478 & 240 \\
Compact GS   & 0:38:03 & 1.28 & 6194 & 22.72 & 0.6716 & 0.3499 & 295 \\
Difix        & 0:19:19 & 1.34 & 3594 & 22.63 & 0.7584 & \cellcolor[HTML]{EA9999}0.2196 & 280 \\
\midrule
Ours         & \cellcolor[HTML]{F9CB9C}0:13:52 & \cellcolor[HTML]{EA9999} 0.84 & \cellcolor[HTML]{F9CB9C}2582 & \cellcolor[HTML]{EA9999}24.45 & \cellcolor[HTML]{EA9999}0.8282 &  0.2222 & \cellcolor[HTML]{F9CB9C}458 \\
Ours+ReLU    & \cellcolor[HTML]{EA9999}0:13:43 & \cellcolor[HTML]{F9CB9C}0.86 & \cellcolor[HTML]{EA9999}2500 & \cellcolor[HTML]{F9CB9C}24.15 & \cellcolor[HTML]{F9CB9C}0.8248 & \cellcolor[HTML]{F9CB9C}0.2210 & \cellcolor[HTML]{EA9999}466 \\
\midrule
\multicolumn{8}{c}{\cellcolor{gray!20}\textit{room}} \\
\midrule
3DGS         & 0:30:08 & 2.78 & 7546 & 26.03 & 0.4692 & 0.5269 & 135 \\
Compact GS   & 0:53:43 & 2.48 & 10778 & 24.77 & 0.4350 & 0.5456 & 147 \\
Difix        & 0:11:41 & 0.48 & 3898  & 28.61 & \cellcolor[HTML]{F9CB9C}0.9133 & \cellcolor[HTML]{EA9999}0.2322 & 820 \\
\midrule
Ours         & \cellcolor[HTML]{F9CB9C}0:10:24 & \cellcolor[HTML]{F9CB9C}0.32 & \cellcolor[HTML]{F9CB9C}3378  & \cellcolor[HTML]{F9CB9C}30.17 & 0.9103 & 0.2691 & \cellcolor[HTML]{F9CB9C}1154 \\
Ours+ReLU    & \cellcolor[HTML]{EA9999}0:09:44 & \cellcolor[HTML]{EA9999}0.31 & \cellcolor[HTML]{EA9999}1758  & \cellcolor[HTML]{EA9999}30.22 & \cellcolor[HTML]{EA9999}0.9133 & \cellcolor[HTML]{F9CB9C}0.2666 & \cellcolor[HTML]{EA9999}1230 \\
\midrule
\multicolumn{8}{c}{\cellcolor{gray!20}\textit{trex}} \\
\midrule
3DGS         & 0:24:29 & 2.11 & 5662 & 24.86 & 0.5936 & 0.4254 & 182 \\
Compact GS   & 0:41:22 & 1.63 & 8118 & 24.13 & 0.5728 & 0.4401 & 223 \\
Difix        & 0:13:31 & 0.79 & \cellcolor[HTML]{F9CB9C}3140  & 24.75 & 0.8571 & 0.2601 & 502 \\
\midrule
Ours         & \cellcolor[HTML]{F9CB9C}0:10:53 & \cellcolor[HTML]{F9CB9C}0.44 & 3666  & \cellcolor[HTML]{F9CB9C}26.64 & \cellcolor[HTML]{F9CB9C}0.8838 & \cellcolor[HTML]{F9CB9C}0.2510 & \cellcolor[HTML]{F9CB9C}895 \\
Ours+ReLU    & \cellcolor[HTML]{EA9999}0:10:30 & \cellcolor[HTML]{EA9999}0.43 & \cellcolor[HTML]{EA9999}2186 & \cellcolor[HTML]{EA9999}26.74 & \cellcolor[HTML]{EA9999}0.8875 & \cellcolor[HTML]{EA9999}0.2464 & \cellcolor[HTML]{EA9999}927 \\

\bottomrule
\end{tabular}}
\label{tab10_1}
\end{table*}

\end{document}